\documentclass[11pt]{article}

\usepackage[preprint]{acl}

\usepackage{times,latexsym}
\usepackage[T1]{fontenc}
\usepackage[utf8]{inputenc}
\usepackage{microtype,graphicx}

\usepackage{booktabs,multirow,subcaption}
\usepackage{tikz}
\usepackage{pgfplots}
\pgfplotsset{compat=1.18}
\graphicspath{{figures/}}

\usepackage{amsmath,amssymb}
\usepackage{enumitem}              %
\usepackage[nameinlink]{cleveref}   %
\usepackage{xspace}
\usepackage[normalem]{ulem}         %

\providecommand{\noopsort}[1]{}
\newcommand{\code}[1]{\texttt{#1}}
\definecolor{tealBg}{HTML}{E0F7F7}
\definecolor{tealObj}{HTML}{086B6C}
\definecolor{tealText}{HTML}{027C7C}
\definecolor{purpleBg}{HTML}{F6EDFF}
\definecolor{purpleObj}{HTML}{9751D7}
\definecolor{purpleText}{HTML}{5B2987}
\definecolor{vermilionBg}{HTML}{FFEDE4}
\definecolor{vermilionObj}{HTML}{CC8463}
\definecolor{vermilionText}{HTML}{743B1F}
\definecolor{grayBg}{HTML}{F2F3F1}
\definecolor{grayObj}{HTML}{868780}
\definecolor{grayText}{HTML}{42433D}
\definecolor{orangeBg}{HTML}{FFF0DB}
\definecolor{orangeObj}{HTML}{E2850C}
\definecolor{orangeText}{HTML}{8F4E00}
\colorlet{currentBg}{tealBg}
\colorlet{currentObj}{tealObj}
\colorlet{currentText}{tealText}
\colorlet{futureBg}{purpleBg}
\colorlet{futureObj}{purpleObj}
\colorlet{futureText}{purpleText}
\colorlet{oursBg}{tealBg}
\colorlet{oursObj}{tealObj}
\colorlet{oursText}{tealText}
\colorlet{negativeBg}{vermilionBg}
\colorlet{negativeObj}{vermilionObj}
\colorlet{negativeText}{vermilionText}
\colorlet{baselineBg}{grayBg}
\colorlet{baselineObj}{grayObj}
\colorlet{baselineText}{grayText}
\colorlet{contextBg}{grayBg}
\colorlet{contextObj}{grayObj}
\colorlet{contextText}{grayText}

\newcommand{\better}[1]{\textcolor{oursText}{#1}}
\newcommand{\worse}[1]{\textcolor{negativeText}{#1}}
\newcommand{\methodname}{\textit{B3}\xspace}

\title{Blurring the Block Boundaries: Overlapping Block Denoising for Efficient Diffusion Language Models}

\author{
  \textbf{Amir Mann\textsuperscript{1}},
  \textbf{Jonathan Fhima\textsuperscript{2}},
  \textbf{Shaked Brody\textsuperscript{2}},
  \textbf{Or Avitan\textsuperscript{2}},
\\
  \textbf{Asif Zamir\textsuperscript{2}},
  \textbf{Shahar Tsiper\textsuperscript{2}},
  \textbf{Niv Nayman\textsuperscript{2}}
}

\begin{document}
\maketitle
\begingroup
\renewcommand{\thefootnote}{}
\footnotetext{\textsuperscript{1}Technion -- Israel Institute of Technology. Work done during an internship at Amazon.
  \textsuperscript{2}Amazon Web Services.
  Correspondence to Jonathan Fhima, \href{mailto:fhimajon@amazon.com}{fhimajon@amazon.com}.
  Code: \url{https://github.com/amazon-science/BlurringBlockBoundaries}}
\endgroup

\begin{abstract}

Diffusion Language Models (DLMs) have emerged as effective alternatives to autoregressive models for latency-critical tasks, enabling faster token generation.
Recent advances employ block diffusion with self-editing capabilities, where each block refines tokens iteratively before advancing to the next one.
Yet each block is only as fast as its slowest token, and the next block cannot start before it.
We suggest Blurring the Block Boundaries (\methodname), a novel method with which two blocks are refined in parallel within a shared forward pass.
\methodname calibrates itself on the fly, opening the next block only once the model is ready to start generating it.
It is a training-free and hyperparameter-free acceleration method which seamlessly integrates within any self-editing block diffusion model.
Evaluated on state-of-the-art DLMs across 10 diverse datasets spanning varied tasks and difficulty levels, \methodname results in on par generation quality and 11.5--14.5\% speedup compared to the base model, pushing the Pareto frontier for latency--quality trade-offs in DLM inference.

\end{abstract}

\section{Introduction}
\label{sec:intro}

\begin{figure}[!t]

  \centering
  \includegraphics{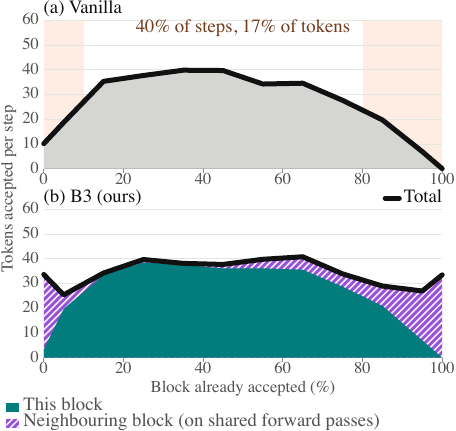}
  \caption[Teaser]{Tokens accepted per denoising step against the share of the block already accepted, for DiffusionGemma 26B on HumanEval+. (a) Vanilla decoding: the orange bands mark low-acceptance regions at the block's boundaries, which take $40\%$ of the steps but yield only $17\%$ of the tokens. (b) \methodname{}: forward passes shared with the neighbouring block (purple) keep the tokens accepted per step consistent across the block.}

  \label{fig:teaser}
\end{figure}
\begin{figure*}[t]

  \centering
  \includegraphics{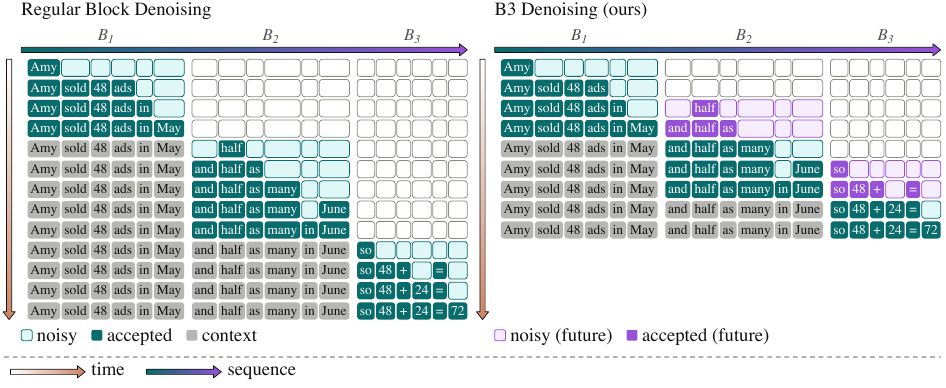}
    \caption{Schematic of \methodname{}. Each row is one forward pass. Left: vanilla decoding denoises one block at a time. Right: \methodname{} starts the next block early, sharing forward passes with the current block, and finishes in fewer passes.
    }
  \label{fig:method}
\end{figure*}

Discrete Diffusion Language Models (DLMs)~\cite{campbell2022ctmc,gat2024dfm,holderrieth2026flowsdiffusions,nie2025llada,zhu2025llada15,ye2025dream7b,bie2025llada2} decode multiple tokens in parallel at each forward pass, offering substantially faster generation than autoregressive (AR) models~\cite{song2025seeddiffusion,diffusiongemma2026,wang2026d2f,man2026dodo}.

Early DLMs denoised the whole output within a single canvas with bidirectional attention~\cite{austin2021d3pm,lou2024sedd}. Since the canvas length is fixed, every forward pass contains all positions regardless of how many tokens need updating, and generation cannot extend beyond the canvas boundary. Block diffusion~\cite{arriola2025blockdiffusion} addresses this by splitting the canvas into blocks decoded sequentially: each forward pass refines only the current block, new blocks can be appended as needed, and each block attends causally to the previous blocks via a KV cache. This approach is widely used in recent large-scale DLMs~\cite{bie2025llada2,wu2026fastdllmv2,cheng2025sdar,song2025seeddiffusion,wang2026d2f}. State-of-the-art block DLMs~\cite{diffusiongemma2026,bie2026llada21} learn to self-edit from uniform-state corruption~\cite{austin2021d3pm}, and may spend further forward passes refining a block before moving on to the next.

A key observation of ours is that this artificial split into blocks carries an inherent inefficiency at block boundaries. Tokens are not equally easy to predict: some settle in a step or two, others need many more, and a block is not finalized until its slowest token is accepted.
A forward pass over a block has the same computational cost regardless of how many tokens are accepted.
At the end of a block's denoising process, the model accepts its last few tokens or self-edits earlier ones, while the next block's easy to predict tokens wait behind them, untouched. We show empirically that the number of tokens accepted per forward pass drops sharply near the block boundaries (\Cref{fig:teaser}(a)), consistently across models and datasets (\Cref{sec:appendix_block}).

In this work we present \textbf{Blurring the Block Boundaries (\methodname)}, a novel approach that reduces this loss in token throughput by blurring the artificial block boundaries. Once the current block is settled enough, the next block is drafted in a shared forward pass, conditioned on the current block's evolving state, parallelizing the end of the current block generation and the beginning of the next one (\Cref{fig:teaser}(b)). Attention stays block-causal, so the current block's predictions are left unchanged. Once the draft becomes the current block, self-editing corrects any errors it carries from being initialised before the previous block had fully converged. The spawning time of the next block is dynamically determined from statistics of the predicted tokens in previous blocks, making \methodname training-free, thus requiring no adaptation to integrate into existing models.

Evaluated on DiffusionGemma 26B~\cite{diffusiongemma2026}, LLaDA 2.1 mini 16B, and LLaDA 2.1 flash 100B~\cite{bie2026llada21} across 10 diverse datasets, \methodname delivers 11.5--14.5\% speedup at on par generation quality without any training or hyperparameter tuning, advancing the latency--quality Pareto frontier of DLM inference.

To summarise, our key contributions are:

\begin{itemize}
    \item Identifying an inefficiency of block diffusion, where the number of tokens accepted per forward pass drops near the block boundaries.
    \item Introducing \methodname{}, a novel approach which denoises two consecutive blocks in shared forward passes.
    \item Designing a self-calibrating spawning criterion that requires no training or hyperparameter tuning.
    \item Extensive evaluation of \methodname{} on three self-editing state-of-the-art block DLMs and 10 benchmarks.
\end{itemize}

\section{Related Work}
\label{sec:related}

\paragraph{Diffusion language models.}
Diffusion language models denoise either continuous token embeddings~\cite{li2022diffusionlm,gulrajani2023plaid,chen2026langflow} or discrete tokens. Discrete diffusion models are characterised by their corruption process, which determines whether generation can revise its own tokens. Masked diffusion replaces tokens with an absorbing \texttt{[MASK]} state~\cite{sahoo2024mdlm,ou2025radd,nie2025llada,ye2025dream7b}, so generation progressively unmasks positions while revealed tokens are not revisited. Uniform-state diffusion replaces them with random vocabulary tokens~\cite{austin2021d3pm,sahoo2025duality,ye2026sumi,diffusiongemma2026}, so every position is re-predicted at every step. Hybrid models mix both~\cite{vonrutte2025gidd,bie2026llada21}. Masked models can be extended with self-editing through remasking~\cite{wang2025remdm} or self-correction objectives~\cite{schiff2026selfcorrecting,wang2026gdd}. Block DLMs~\cite{arriola2025blockdiffusion}, adapted from pretrained AR models~\cite{wu2026fastdllmv2,fu2026efficientdlm} or scaled to 100B parameters~\cite{bie2025llada2}, now underlie most recent large-scale DLMs.

\paragraph{Accelerating DLM inference.}
Block diffusion~\cite{arriola2025blockdiffusion} decodes the sequence block by block, so the clean prefix is cached exactly rather than approximated as in full-canvas DLMs~\cite{wu2026fastdllm,ma2025dkvcache,hu2026flashdlm,liu2026dllmcache,jiang2026d2cache}. Within each block, sampling strategies accept more tokens per step via confidence thresholds~\cite{wu2026fastdllm,bie2025llada2}, entropy bounds~\cite{benhamu2025eb} and adaptive parallel decoding~\cite{israel2025}.

\paragraph{Multi-block decoding.}
Orthogonally to these accelerations, multi-block methods change which blocks share a forward pass, and therefore run on top of the model's native sampler. For DLMs trained without blocks, sliding and adaptive schemes reshape the block imposed by the sampler at inference~\cite{luo2026dsb,zhang2026swordsman,lu2026adablock,wang2026whentocommit,wan2026geoblock,zou2026ahd}. Concurrently to our work, FlowBlock~\cite{tian2026flowblock} and MBD-LMs~\cite{jin2026multiblock} overlap consecutive blocks of block DLMs, advancing to the next block at a tuned progress threshold. In contrast, \methodname{} needs no per-model or per-task tuning and achieves a better latency--quality trade-off than both (\Cref{tab:baselines_llada}). \methodname{} also applies to USDMs, to which others are not adapted.
\section{Method}
\label{sec:method}
\begin{figure}[tbp]

  \centering
  \includegraphics{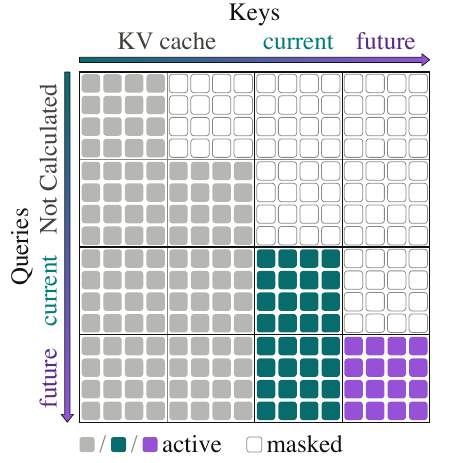}
  \caption{\methodname{} attention mask of a shared forward pass; the current block never attends to the future block.}
  \label{fig:attention}
\end{figure}

\subsection{Preliminaries}

Discrete diffusion language models generate a sequence $x=(x^1,\dots,x^L)$ by learning to reverse a token-level corruption process. At noise level $t\in[0,1]$, from clean ($t=0$) to fully corrupted ($t=1$), each position $\ell\in\{1,\dots,L\}$ of the noisy sequence $x_t$ independently either keeps token $x^\ell_t$ or is replaced by a token drawn from a corruption distribution $\pi$, with the replacement probability increasing in $t$. A denoiser $p_\theta(x^\ell \mid x_t)$ predicts the clean token at every position $\ell$. For MDMs, $\pi$ always yields a special \texttt{[MASK]} token. For USDMs, $\pi$ is uniform over the vocabulary, so any token may be corrupted and $p_\theta$ learns to re-predict every position, enabling self-editing. Hybrid models~\cite{bie2026llada21, vonrutte2025gidd} mix the two corruption distributions at train time, and so inherit the self-editing capabilities of the USDM component.

At inference, every position is initialised from $\pi$ and $p_\theta$ is applied iteratively. Each denoising step is one forward pass of $p_\theta$ that predicts tokens at all $L$ positions. A sampler selects a subset of these predicted tokens to retain in the sequence for the next denoising step by using a confidence threshold~\cite{wu2026fastdllm,bie2025llada2} or an entropy bound~\cite{benhamu2025eb}. 
Retained tokens accumulate over denoising steps until the whole sequence is complete.

Block diffusion models~\cite{arriola2025blockdiffusion} split the sequence into bidirectional blocks and generate them causally. We write $x^{b}$ for the tokens in the $b$-th block $B_b\subset\{1,\dots,L\}$, $x^{<b}$ for tokens in all preceding blocks, and $\ell\in B_b$ for a position $\ell$ in block $b$. The sampler runs only within the current block $b$, and each step denoises the block's noisy state $x^b_{t_b}$ at noise level $t_b$, conditioned on the clean prefix:
\begin{equation}
    p_\theta\big(x^b \mid x^{<b}, x^b_{t_b}\big)
    \label{eq:vanilla_forward}
\end{equation}

Once $x^b$ meets the sampler's convergence criterion, $x^b$ is finalized and moved into the context, and block $b+1$ is initialised from $\pi$, unless an EOS has been generated in $x^b$. Attention is block-causal: a position $\ell\in B_b$ attends only to blocks $1,\dots,b$, so each forward pass processes only block $b$, reading $x^{<b}$ from an exact KV cache.

\subsection{Blurring the Block Boundaries}

As block diffusion turns every block boundary into a synchronisation barrier, no position in block $b+1$ may start until the entire block $b$ has been finalized. Empirical analysis (\Cref{sec:boundary}) identifies a significant efficiency drop in token generation around those artificial boundaries.

Rather than waiting for the current block to be finalized before initialising the next one, \methodname{} starts denoising the future block simultaneously with the current block (\Cref{fig:method}) when approaching the boundaries: once the current block $b$ is \emph{settled enough} according to our proposed criterion (\Cref{sec:method_spawning}), the future block $b+1$ is initialised and both blocks are denoised together until $b$ is finalized.
Concretely, both blocks share a single forward pass through $p_\theta$:
\begin{equation}
    p_\theta\big(x^b, x^{b+1} \mid x^{<b}, x^b_{t_b}, x^{b+1}_{t_{b+1}}\big)
    \label{eq:joint_forward}
\end{equation}

Because attention is block-causal (\Cref{fig:attention}), no position in block $b$ attends to $x^{b+1}_{t_{b+1}}$, so the predictions for $x^b$ are identical to those of the vanilla forward pass of \cref{eq:vanilla_forward}.

Each block keeps its own sampler, with separate noise levels $t_b$ and $t_{b+1}$, so block $b$ is denoised exactly as in standard block decoding.
The future block, however, is conditioned on $x^b_{t_b}$, which may still contain tokens sampled from $\pi$, that can mislead its early predictions. Self-editing lets the sampler of block $b+1$ revise already accepted tokens of $x^{b+1}$ after $x^b$ converges. Once $x^b$ is finalized, block $b+1$ becomes the current block and continues its denoising conditioned on the clean prefix $x^{<b+1}$.

\subsection{Future Block Spawning}
\label{sec:method_spawning}

\begin{table*}[tp]

  \centering
  \small
  \setlength{\tabcolsep}{5pt}
  \begin{tabular}{ll r rrr rrr rrr}
    \toprule
     & & & \multicolumn{3}{c}{TPS} & \multicolumn{3}{c}{TPF} & \multicolumn{3}{c}{Accuracy} \\
    \cmidrule(lr){4-6} \cmidrule(lr){7-9} \cmidrule(lr){10-12}
    Model & Benchmark & $n$ & Base & \methodname (ours) & $\Delta\%$ & Base & \methodname (ours) & $\Delta\%$ & Base & \methodname (ours) & $\Delta pp$ \\
    \midrule
    \multirow{11}{*}{\rotatebox{90}{DiffusionGemma 26B}}
    & GSM8K & 1319 & 136.1 & 150.8 & \better{+10.8} & 28.74 & 33.69 & \better{+17.2} & 95.75 & 96.13 & \better{+0.38} \\
    & MATH500 & 500 & 139.9 & 156.0 & \better{+11.6} & 28.69 & 34.16 & \better{+19.1} & 95.80 & 92.40 & \worse{-3.40} \\
    & AIME26 & 30 & 101.0 & 121.6 & \better{+20.3} & 20.61 & 25.48 & \better{+23.6} & 73.33 & 76.67 & \better{+3.33} \\
    & Omni-MATH & 4428 & 107.6 & 118.0 & \better{+9.6} & 21.70 & 25.10 & \better{+15.7} & 61.50 & 60.68 & \worse{-0.81} \\
    & GPQA-D & 198 & 94.4 & 107.1 & \better{+13.4} & 19.18 & 22.88 & \better{+19.3} & 73.74 & 71.72 & \worse{-2.02} \\
    & IFEval & 541 & 103.4 & 119.9 & \better{+16.0} & 21.08 & 26.61 & \better{+26.3} & 95.80 & 95.44 & \worse{-0.36} \\
    & HumanEval+ & 164 & 148.9 & 160.7 & \better{+7.9} & 30.77 & 37.53 & \better{+22.0} & 91.46 & 89.63 & \worse{-1.83} \\
    & MBPP+ & 378 & 136.4 & 149.6 & \better{+9.7} & 28.39 & 33.40 & \better{+17.6} & 80.69 & 82.28 & \better{+1.59} \\
    & BBEH & 4520 & 116.2 & 126.6 & \better{+8.9} & 24.16 & 28.53 & \better{+18.1} & 36.72 & 37.01 & \better{+0.29} \\
    & LCB & 342 & 108.0 & 119.2 & \better{+10.4} & 22.11 & 25.59 & \better{+15.7} & 70.47 & 69.30 & \worse{-1.17} \\
    \cmidrule(lr){2-12}
    & Average &  & 119.2 & 132.9 & \better{+11.5} & 24.54 & 29.30 & \better{+19.4} & 77.53 & 77.13 & \worse{-0.40} \\
    \midrule
    \multirow{11}{*}{\rotatebox{90}{LLaDA 2.1 flash 100B}}
    & GSM8K & 1319 & 90.7 & 98.5 & \better{+8.5} & 5.58 & 6.37 & \better{+14.0} & 94.01 & 93.56 & \worse{-0.45} \\
    & MATH500 & 500 & 106.6 & 123.1 & \better{+15.5} & 6.56 & 7.55 & \better{+15.1} & 95.20 & 94.60 & \worse{-0.60} \\
    & AIME26 & 30 & 75.4 & 82.0 & \better{+8.7} & 4.71 & 5.21 & \better{+10.4} & 66.67 & 76.67 & \better{+10.00} \\
    & Omni-MATH & 4428 & 77.2 & 84.9 & \better{+10.1} & 4.73 & 5.24 & \better{+10.8} & 59.94 & 59.49 & \worse{-0.45} \\
    & GPQA-D & 198 & 52.9 & 53.6 & \better{+1.2} & 3.25 & 3.43 & \better{+5.7} & 66.67 & 68.18 & \better{+1.52} \\
    & IFEval & 541 & 26.2 & 27.8 & \better{+5.9} & 1.62 & 1.71 & \better{+5.9} & 88.85 & 89.21 & \better{+0.36} \\
    & HumanEval+ & 164 & 211.1 & 254.3 & \better{+20.4} & 13.20 & 16.40 & \better{+24.2} & 85.37 & 84.76 & \worse{-0.61} \\
    & MBPP+ & 378 & 131.7 & 157.5 & \better{+19.6} & 8.16 & 9.90 & \better{+21.4} & 76.72 & 74.34 & \worse{-2.38} \\
    & BBEH & 4520 & 66.2 & 71.6 & \better{+8.2} & 4.07 & 4.48 & \better{+10.2} & 12.48 & 14.16 & \better{+1.67} \\
    & LCB & 342 & 87.9 & 98.2 & \better{+11.7} & 5.49 & 6.21 & \better{+13.1} & 41.52 & 40.06 & \worse{-1.46} \\
    \cmidrule(lr){2-12}
    & Average &  & 92.6 & 105.2 & \better{+13.6} & 5.74 & 6.65 & \better{+15.9} & 68.74 & 69.50 & \better{+0.76} \\
    \midrule
    \multirow{11}{*}{\rotatebox{90}{LLaDA 2.1 mini 16B}}
    & GSM8K & 1319 & 114.2 & 124.8 & \better{+9.3} & 4.88 & 5.49 & \better{+12.5} & 90.90 & 89.84 & \worse{-1.06} \\
    & MATH500 & 500 & 160.4 & 186.1 & \better{+16.0} & 6.85 & 8.14 & \better{+18.9} & 92.80 & 91.80 & \worse{-1.00} \\
    & AIME26 & 30 & 113.7 & 125.9 & \better{+10.7} & 4.92 & 5.46 & \better{+11.0} & 46.67 & 50.00 & \better{+3.33} \\
    & Omni-MATH & 4428 & 115.5 & 127.8 & \better{+10.7} & 4.91 & 5.55 & \better{+13.0} & 46.39 & 45.93 & \worse{-0.45} \\
    & GPQA-D & 198 & 69.5 & 73.8 & \better{+6.1} & 3.02 & 3.26 & \better{+8.0} & 52.53 & 52.02 & \worse{-0.51} \\
    & IFEval & 541 & 31.1 & 32.4 & \better{+4.2} & 1.35 & 1.40 & \better{+4.3} & 86.45 & 85.97 & \worse{-0.48} \\
    & HumanEval+ & 164 & 247.2 & 301.4 & \better{+21.9} & 10.94 & 13.69 & \better{+25.2} & 83.54 & 81.71 & \worse{-1.83} \\
    & MBPP+ & 378 & 163.6 & 194.2 & \better{+18.7} & 7.16 & 8.57 & \better{+19.6} & 69.84 & 69.05 & \worse{-0.79} \\
    & BBEH & 4520 & 69.4 & 73.5 & \better{+5.8} & 2.98 & 3.22 & \better{+7.8} & 6.91 & 5.94 & \worse{-0.96} \\
    & LCB & 342 & 113.7 & 131.8 & \better{+15.9} & 5.10 & 5.83 & \better{+14.3} & 25.73 & 26.61 & \better{+0.88} \\
    \cmidrule(lr){2-12}
    & Average &  & 119.8 & 137.2 & \better{+14.5} & 5.21 & 6.06 & \better{+16.3} & 60.17 & 59.89 & \worse{-0.28} \\
    \bottomrule
  \end{tabular}

\caption{Per-benchmark results for \methodname{} on DiffusionGemma 26B, LLaDA 2.1 flash 100B, and LLaDA 2.1 mini 16B. 
}
  \label{tab:results}

\end{table*}

During training, different blocks are denoised disjointly.
On the one hand, waiting for the current block to fully converge at inference time spawns the next block too late and prevents the desired overlap. On the other hand, spawning the future block too early conditions it on a still evolving current block, which may cause inference to deviate too far from training. Hence, the future block should be spawned once the tokens of the current block are settled enough. 

The model entropy is a well-established~\cite{benhamu2025eb, zhang2026swordsman} measure of how settled a token is,
\begin{align}
  H^\ell_i = -\sum_{v\in\mathcal{V}} &p_\theta\big(x^\ell=v \mid x^{<b}, x^{b}_{t_{b}}\big)\cdot \\
  &\log p_\theta\big(x^\ell=v \mid x^{<b}, x^{b}_{t_{b}}\big) \nonumber
  \label{eq:entropy}
\end{align}
at position $\ell\in B_{b}$ on forward pass $i$ of the denoising process at the corresponding noise level $t_b$, with the vocabulary set $\mathcal{V}$.
The future block is then spawned once the entropy of every token in the current block $b$ is at most a threshold $\tau$,
\begin{equation}
  \max_{\ell \in B_b} H^\ell_i \leq \tau
  \label{eq:gate}
\end{equation}
For the method to remain hyperparameter-free, the remaining question is how to set $\tau$ without tuning it per model or task.

Intuitively, \cref{eq:gate} asks whether the tokens of $B_b$ are settled enough for the model to start accepting tokens in $B_{b+1}$.
Since $B_{b+1}$ attends to $B_b$ block-causally, we look at the steps on which the model accepts a new rightmost token in $B_b$.
Accepting a new rightmost token is similar to accepting the first token of $B_{b+1}$, because all tokens to its right are still unsettled.
The entropy levels of these steps set $\tau$ on the model's own scale.
On such a step, the positions to the left of the newly accepted token play the role of $B_b$ in \cref{eq:gate}, and the new token plays the role of $B_{b+1}$.
The analogy needs two properties, (i) such steps occur gradually through the block, and (ii) the new token is accepted mainly on the basis of the partly settled tokens to its left.
Both are detailed and supported in \Cref{sec:appendix_ltr}.

\paragraph{Generation frontier.}
Let $I_i \subseteq \{1,\dots,L\}$ be the set of positions accepted over the generation up to forward pass $i$, accumulated across blocks.\footnote{For DiffusionGemma, see \Cref{sec:appendix_gemma}.}
We call its rightmost position, $r_i = \max I_i$, the generation frontier.
When the frontier advances, $r_i > r_{i-1}$, the newly accepted token at $r_i$ has all of its accepted context to its left, like the first token of $B_{b+1}$.

\paragraph{Threshold candidates.}
Each advance of the frontier shows an entropy level at which the model was ready to accept the next token.
That level is the highest entropy among the positions to the left of $r_i$ on the previous pass.
This level is therefore a candidate for $\tau$:
\begin{equation}
  \tau^{\text{candidate}}_{i} = \max_{\ell \in B_b,\; \ell < r_{i}} H^\ell_{i-1}
  \label{eq:left_gate}
\end{equation}
\Cref{eq:left_gate} also matches \cref{eq:gate} in two details.
First, $r_i$ itself is excluded, just as the first token of $B_{b+1}$ is excluded from \cref{eq:gate}.
Second, on pass $i$ the gate decides from the entropies of pass $i$ whether tokens of $B_{b+1}$ may be accepted from pass $i+1$ on, so \cref{eq:left_gate} introduces the same one-step lag by using the entropies of pass $i-1$.

\paragraph{Threshold.}
Candidates are collected over the generation at hand, on the forward passes on which the frontier advances.
Passes with an active future block are excluded, because the frontier then lies in $B_{b+1}$ and not in $B_b$.
This gives
\[
  C_\tau=\{i \mid r_i > r_{i-1},\; B_{b+1} \text{ is not active at } i\}.
\]
Averaging the candidates of $C_\tau$ gives the threshold,
\begin{equation}
  \tau = \frac{1}{|C_\tau|} \sum_{i \in C_\tau} \tau^{\text{candidate}}_{i} .
  \label{eq:tau_calc}
\end{equation}
$\tau$ is updated as candidates arrive, and until the first candidate exists, the next block is spawned only once the current block is fully accepted.
Finally, to avoid spawning future blocks past the end of the sequence, spawning is also gated on the current block not predicting an EOS.
\section{Experiments}
\label{sec:experiments}
\begin{figure*}[tp]

  \centering
  \includegraphics{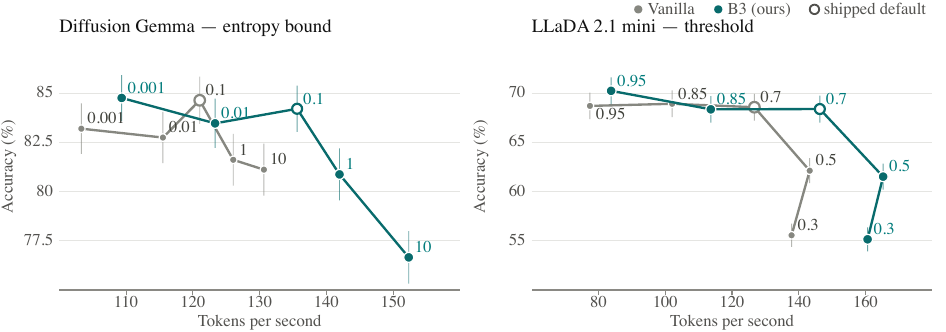}
  \caption{Sweep of the acceptance criterion. Error bars denote $\pm 1$ standard error of the average accuracy due to question sampling, combined from per-benchmark binomial standard errors.
  }
  \label{fig:pareto}
\end{figure*}
\begin{figure*}[t]
  \centering
  \includegraphics{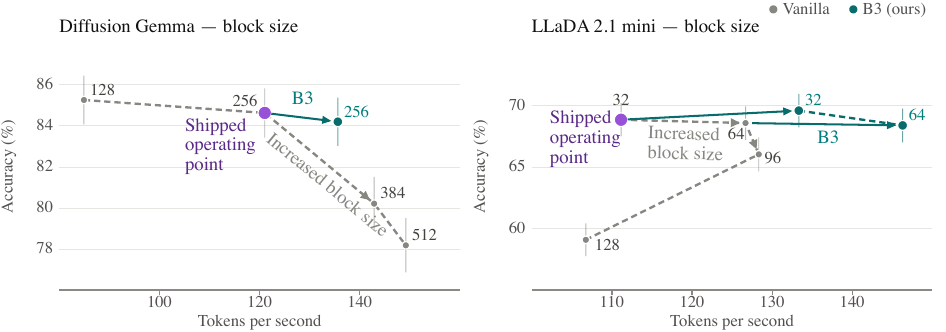}
  \caption{\methodname{} shows a preferable accuracy/speed trade-off compared to increasing the block size. At each shipped operating point, it is better to use \methodname{} instead of increasing the block size.}
  \label{fig:ablation_block}

\end{figure*}

\paragraph{Models.}
We evaluate \methodname{} on DiffusionGemma 26B~\cite{diffusiongemma2026}, LLaDA 2.1 flash 100B and LLaDA 2.1 mini 16B~\cite{bie2026llada21}, using their public checkpoints and Hugging Face implementations. DiffusionGemma uses its shipped decoding defaults. For both LLaDA variants we use block size 64 instead of the default 32, as it is faster at comparable accuracy (\Cref{fig:ablation_block}). This choice favors the baseline, as it lowers the relative speedup of \methodname{} from 20\% to 15\%. Hardware and other technical details in \Cref{sec:appendix_technical}.

\paragraph{Metrics.}
We report Tokens per Second (TPS) calculated as output tokens divided by generation time in seconds, and Tokens per Forward (TPF) calculated as output tokens divided by model forward passes. We compute TPS and TPF for every input sample and report the mean over the benchmark. $\Delta\%$ compares these means, $(T_{method}-T_{base})/T_{base}$. In the Average row, we first average TPS (or TPF) over benchmarks and then take $\Delta\%$. For a measure of quality we report Accuracy, defined as is common per dataset, and for ease of comparison we report also $\Delta pp$ measuring the difference in absolute performance (accuracy is already scaled as a unitless fraction) $\Delta pp = A_{method}-A_{base}$. For all relative metrics we mark in \better{teal} improvements over the baseline and in \worse{vermilion} degradation over it. 

\paragraph{Datasets.} We use the full test sets of 10 benchmarks, zero-shot except for 8-shot GSM8K (Using \code{gsm8k\_cot} of \code{lm-evaluation-harness};~\citealp{gao2024evalharness}), all with their common metrics.

\begin{itemize}[leftmargin=*, labelsep=0.4em, nosep]
  \item Math: GSM8K~\cite{cobbe2021gsm8k}, MATH500~\cite{hendrycks2021math,lightman2024verify}, AIME 2026~\cite{balunovic2025matharena} and Omni-MATH~\cite{gao2025omnimath}, scored by exact match of the final answer, checked for mathematical equivalence with Hugging Face Math-Verify\footnote{\url{https://github.com/huggingface/Math-Verify}} on MATH500 and Omni-MATH.
  \item Science: GPQA Diamond~\cite{rein2024gpqa}.
  \item Reasoning: BBEH~\cite{kazemi2025bbeh}, scored by adjusted harmonic mean of task accuracy.
  \item Instruction following: IFEval~\cite{zhou2023ifeval}, loose instruction-level accuracy.
  \item Code: HumanEval+ and MBPP+~\cite{liu2023evalplus} and LiveCodeBench~\cite{jain2025livecodebench}, scored as pass@1.
\end{itemize}

Since \methodname{} needs no per-model or per-task tuning, no validation sets are used. Due to compute
constraints, ablations and multi-block comparisons (\Cref{sec:comparison,sec:speed-acc})
exclude Omni-MATH and BBEH (4428 and 4520 examples) and report averages
over the remaining eight, referred to as the \emph{light} benchmarks.

\subsection{Results}

\Cref{tab:results} reports results on DiffusionGemma, LLaDA 2.1 flash and LLaDA 2.1 mini, spanning two model families and 16B to 100B parameters. 
Averaged over benchmarks, \methodname{} raises TPS by 11.5--14.5\% and TPF by 15.9--19.4\%, while
average accuracy changes only by --0.40 to +0.76\,pp. 
The acceleration is consistent: \methodname{} improves both TPS and TPF on every one of the 30 model--benchmark pairs, without any per-model or per-task tuning.
All per-benchmark accuracy changes lie within their 95\% intervals, over three seeds for DiffusionGemma and by a paired bootstrap over problems for the greedy LLaDA~2.1 mini (\Cref{tab:appendix_uncertainty}).
DiffusionGemma gains on five of the eight light benchmarks, while LLaDA~2.1 mini loses on six, the largest being HumanEval+ ($-1.8$\,pp) and GSM8K ($-1.1$\,pp).
MATH500 is a real cost on DiffusionGemma: its $-3.4$\,pp drop is $-2.3$\,pp over seeds and negative on every seed.
The AIME gains are within noise, and the $+10.0$\,pp on LLaDA~2.1 flash is a net 3 of 30 problems.

\subsection{Comparison with Other Multi-Block Methods}
\label{sec:comparison}
\Cref{tab:baselines_llada} compares \methodname{} with FlowBlock and MBD-LMs (training-free variant) on
LLaDA~2.1 mini, using each method's published hyperparameters on a per-dataset
basis.\footnote{The baselines' math hyperparameters are used for math tasks and GPQA-D, and their code hyperparameters for code tasks and IFEval.
Where the released code differs from the paper, the code is followed.}
Averaged over the eight light benchmarks at LLaDA~2.1's default block size of 32, \methodname{} is the fastest method, raising TPS by $2\%$ more than both baselines, while being $1.7$\,pp and $2.1$\,pp more accurate than FlowBlock and MBD-LMs, respectively.
At the faster block size of 64, \methodname{} again leads on both axes: it is $11\%$ faster than FlowBlock at comparable accuracy, and $3.0$\,pp more accurate than MBD-LMs at $3\%$ higher TPS.
FlowBlock is faster than \methodname{} on every math and science benchmark, while compromising accuracy, so the average speed lead of \methodname{} comes from instruction following and code.
Per-dataset results are in \Cref{tab:appendix_baselines_llada}.

\begin{table}[!t]

  \centering
  \footnotesize
  \setlength{\tabcolsep}{0pt}
  \begin{tabular*}{\columnwidth}{@{\extracolsep{\fill}}lllll@{}}
    \toprule
    Block Size & Method & TPS & TPF & Acc \\
    \midrule
    32 & Base & 111 & 4.83 & 68.8 \\
     & FlowBlock & 132 \better{\scriptsize(+18\%)} & 5.77 \better{\scriptsize(+20\%)} & 67.9 \worse{\scriptsize(-1.0)} \\
     & MBD-LMs & 131 \better{\scriptsize(+18\%)} & \textbf{5.95} \better{\scriptsize(+23\%)} & 67.4 \worse{\scriptsize(-1.4)} \\
     & \textbf{\methodname} & \textbf{133} \better{\scriptsize(+20\%)} & 5.88 \better{\scriptsize(+22\%)} & \textbf{69.6} \better{\scriptsize(+0.7)} \\
    \midrule
    64 & Base & 127 & 5.53 & 68.6 \\
     & FlowBlock & 132 \better{\scriptsize(+4\%)} & 5.89 \better{\scriptsize(+7\%)} & 68.3 \worse{\scriptsize(-0.3)} \\
     & MBD-LMs & 142 \better{\scriptsize(+12\%)} & 6.40 \better{\scriptsize(+16\%)} & 65.3 \worse{\scriptsize(-3.2)} \\
     & \textbf{\methodname} & \textbf{146} \better{\scriptsize(+15\%)} & \textbf{6.48} \better{\scriptsize(+17\%)} & \textbf{68.4} \worse{\scriptsize(-0.2)} \\
    \bottomrule
  \end{tabular*}
  \caption{LLaDA~2.1 mini 16B: \methodname{} vs.\ concurrent inter-block methods on the eight light benchmarks described in \Cref{sec:comparison}.
  }
  \label{tab:baselines_llada}

\end{table}

\subsection{Latency--Quality Trade-off}
\label{sec:speed-acc}
Two axes control the latency--quality trade-off in block diffusion models: the sampler acceptance
criterion and the block size. We ablate both with and without \methodname{}, using
each baseline's native parameters (entropy bound for DiffusionGemma and confidence threshold for
LLaDA~2.1) as the control variable. LLaDA~2.1 is evaluated using the mini model rather than flash
due to compute constraints; full per-benchmark results are shown in \Cref{tab:appendix_block_gemma,tab:appendix_block_llada}.

\paragraph{Sampler threshold.}
\Cref{fig:pareto} sweeps the acceptance criterion with and without \methodname{}. In the high-quality regime, \methodname{} raises TPS at
comparable quality, shifting the frontier, for example gaining +11\% TPS at --0.2\,pp for LLaDA~2.1 mini at threshold 0.95. At the shipped
defaults, \methodname{} is faster than every baseline configuration in the sweep by 3.8\% over the fastest DiffusionGemma baseline, which is 3.1\,pp less accurate, and by 4.3\% over the fastest LLaDA baseline, which is 10.3\,pp less accurate.

\paragraph{Block size.}
\Cref{fig:ablation_block} varies the block size with
and without \methodname{}. For DiffusionGemma, \methodname{} raises TPS by 12\% at block size 256, a
speed the baseline matches only at 384 or larger, at a cost of at least 4.0\,pp quality, while \methodname{} itself loses only 0.4\,pp. For
LLaDA~2.1 mini, blocks beyond 64 give negligible speedup and lose quality, while \methodname{} raises
TPS by 20\% at block size 32 and 15\% at 64 with comparable quality, the latter being faster than
every baseline configuration. Enlarging the block therefore does not recover the speedup of
\methodname{} at comparable quality.

\subsection{Ablation on Future Block Spawning Criteria}
\begin{figure}[t]

  \centering
  \includegraphics{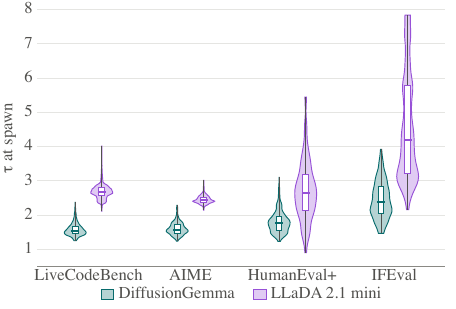}
  \caption{$\tau$ selected by the self-calibrating gate at each spawn, per benchmark and model. Violin: P1--P99; box: P25--P75; line: median.}
  \label{fig:tau_violins}

\end{figure}

\begin{figure}[!t]

  \centering
  \includegraphics{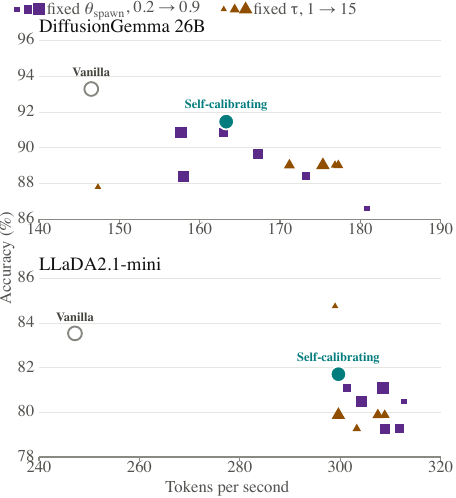}
  \caption{Ablation on HumanEval+ for our self-calibrating spawning threshold vs. spawning at a fixed
  fraction of acceptance, $\theta_{\text{spawn}}$, and at a fixed max entropy gate of $\tau$. Top right is
  preferable.}
  \label{fig:spawn_tau_ablation}

\end{figure}

\begin{figure}[!t]
  \centering
  \includegraphics{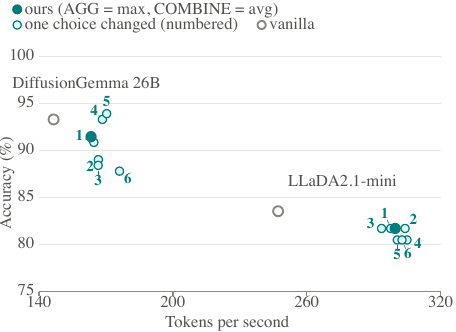}
  \caption{Design choices of the spawn gate on HumanEval+. Each numbered point changes one choice of
    ours (AGG\,=\,max, COMBINE\,=\,avg, $C_\tau$ carried across blocks): (1)~$C_\tau$ resets every block,
    (2)~AGG\,=\,avg, (3)~AGG\,=\,sum, (4)~COMBINE\,=\,avg\,+\,std, (5)~COMBINE\,=\,max, (6)~COMBINE\,=\,max
    (corrected). AGG and COMBINE are defined in \Cref{sec:spawn_ablation}.}
  \label{fig:gate_ablation}

\end{figure}

\label{sec:spawn_ablation}
\Cref{fig:spawn_tau_ablation,fig:gate_ablation} ablate the spawn gate on HumanEval+. In \Cref{fig:spawn_tau_ablation}, each curve sweeps one fixed rule, a fixed accepted fraction $\theta_{\text{spawn}}$ from 0.2 to 0.9 and a fixed $\tau$ from 1 to 15, and our gate is a single point. On DiffusionGemma, a later fixed fraction tends to be slower and more accurate. On LLaDA~2.1 mini, the fixed fractions all reach similar speed and accuracy. Fractions up to 0.9 are less accurate than our gate on both models.

A fixed entropy threshold $\tau$ is no easier to set. The $\tau$ selected by our gate varies across benchmarks and models (\Cref{fig:tau_violins}; see \Cref{sec:appendix_spawn} for a breakdown by dataset and block index). Accuracy is also sensitive to the fixed value chosen (\Cref{fig:spawn_tau_ablation}): $\tau = 1$ beats our gate on LLaDA~2.1 mini but is the least accurate fixed setting on DiffusionGemma. A fixed $\tau$ would therefore need tuning per model and dataset, which \methodname{} avoids.

Finally, \Cref{fig:gate_ablation} changes one design choice of the gate at a time. This includes how the entropies left of $r_i$ are reduced to a candidate (AGG; ours: max, \cref{eq:gate,eq:left_gate}) and how the candidates are combined into $\tau$ (COMBINE; ours: mean, \cref{eq:tau_calc}). All variants land close to ours in speed and accuracy, and none improves on it consistently across both models.

\subsection{Inefficiency at the Block Boundary}
\label{sec:boundary}
\Cref{fig:teaser} illustrates the slow finish of a block under vanilla decoding: acceptance falls as the block nears completion. For DiffusionGemma on HumanEval+, the last 20\% of a block takes 22\% of the steps but yields only 10\% of the tokens (first and last block excluded). This holds for both models on HumanEval+ and GPQA Diamond. The ratio of token share to step share in the last 20\% (1 being parity) is 0.35--0.71 for DiffusionGemma and 0.05--0.62 for LLaDA 2.1 mini (\Cref{fig:end_ratio}). With \methodname{}, the next block accepts tokens on the same forward passes, raising the ratio to 0.67--1.39 and 0.71--1.25. \Cref{sec:appendix_block} breaks this down by block index for DiffusionGemma.
\section{Conclusion}
\label{sec:conclusion}

We introduced \methodname{}, a training-free and without hyperparameters acceleration method for self-editing block diffusion language models. \methodname{} exploits a structural property of these models: late denoising steps of a block accept few tokens, so the next block can start early on the same forward passes. An entropy-based gate calibrates the spawning point from the model's own acceptance behaviour.

Experiments on three models spanning two diffusion families, hybrid (LLaDA 2.1 mini 16B and flash 100B) and USDM (DiffusionGemma 26B), and 10 diverse benchmarks show that \methodname{} consistently improves TPS by 11.5--14.5\% and TPF by 15.9--19.4\%, with an average accuracy change of +0.03\,pp over the three models, without any per-model or per-task tuning. On LLaDA 2.1 mini, \methodname{}, as a plug-and-play addition, also achieves a better latency--quality trade-off than concurrent multi-block methods with tuned thresholds.

Spawning the next block too late forfeits the speedup, while spawning it too early conditions it on a still evolving prefix that can mislead its predictions.
The TPS gains suggest that the gate opens early enough to exploit the low-acceptance tail of each block.
The preserved accuracy suggests that it also opens late enough for the settled tokens of the current block to provide reliable context for the next one.
If any errors do slip past the gate, self-editing appears to correct them.
\section*{Limitations}

\methodname{} relies on self-editing for having low-acceptance forwards to exploit, as well as to correct future block tokens drafted on an unfinished prefix, so masked DLMs without an editing mechanism fall outside its scope. The state-of-the-art models we evaluate, in both diffusion families, self-edit by design. The spawning gate also assumes semi left-to-right generation at parts of the generation, which both evaluated models exhibit (\Cref{sec:appendix_ltr}).

Unlike verification-based speculative decoding, \methodname{} does not guarantee outputs identical to vanilla decoding, although average accuracy changes by only $-0.40$ to $+0.76$\,pp. Its speedup is bounded by the low-acceptance tail of each block. The resulting $11.5$--$14.5\%$ TPS gain is moderate.

\methodname{} aims at latency at small batch sizes, not throughput in large-batch serving. Its wall-clock speedups are measured on one inference stack per model. Due to compute constraints, ablations and comparisons with concurrent methods use LLaDA~2.1 mini and a subset of benchmarks. The evaluation focuses on English only text only tasks.

\appendix
\section{Hardware and Technical Implementation Details}
\label[appendix]{sec:appendix_technical}
\paragraph{Hardware.}
DiffusionGemma runs on 2 A100 40GB GPUs, LLaDA 2.1 flash on 2 H200 141GB GPUs, and LLaDA 2.1 mini on 1 A100 40GB GPU. Generation is capped only by each model's recommended context size: 32K total tokens for DiffusionGemma and 16K output tokens for LLaDA 2.1.

\paragraph{Inference implementation for LLaDA 2.1.}
We fix two inefficiencies in LLaDA 2.1's shipped Hugging Face inference code. First, we replace the MoE forward pass with the fused MoE kernel of vLLM~\cite{kwon2023vllm}, as the public implementation uses a plain for loop to iterate experts, reducing GPU utilization to near 0 in our setup.
We also implement KV caching of finalized blocks. This does not affect the mini model generation speed but without it runtimes for the flash checkpoint are quadratic in length.
Both these changes were unrelated to our method, and we report all results (ours and baselines) over these changes.

\section{Per-Benchmark Results}
\label[appendix]{sec:appendix_per_benchmark}

We detail per-benchmark performance, for the tables and figures which present only averages in \Cref{tab:appendix_pareto_gemma,tab:appendix_pareto_llada,tab:appendix_block_gemma,tab:appendix_block_llada,tab:appendix_baselines_llada}.

\begin{table*}[t]
  \centering
  \footnotesize
  \setlength{\tabcolsep}{0pt}
  \begin{tabular*}{\textwidth}{@{\extracolsep{\fill}}lrrrrrrrrrr@{}}
    \toprule
    Entropy bound & \multicolumn{2}{c}{0.001} & \multicolumn{2}{c}{0.01} & \multicolumn{2}{c}{0.1} & \multicolumn{2}{c}{1} & \multicolumn{2}{c}{10} \\
    \cmidrule(lr){2-3} \cmidrule(lr){4-5} \cmidrule(lr){6-7} \cmidrule(lr){8-9} \cmidrule(lr){10-11}
    Method & Base & \textbf{\methodname} & Base & \textbf{\methodname} & Base & \textbf{\methodname} & Base & \textbf{\methodname} & Base & \textbf{\methodname} \\
    \midrule
    \multicolumn{11}{@{}l}{\textit{TPS}} \\
    GSM8K & 113.5 & \textbf{119.2} & 131.1 & \textbf{141.0} & 136.1 & \textbf{150.8} & 142.3 & \textbf{160.6} & 145.2 & \textbf{167.7} \\
    MATH500 & 125.2 & \textbf{125.1} & 137.3 & \textbf{142.9} & 139.9 & \textbf{156.0} & 144.0 & \textbf{161.2} & 147.0 & \textbf{171.4} \\
    AIME26 & 93.8 & \textbf{97.7} & 99.9 & \textbf{108.2} & 101.0 & \textbf{121.6} & 114.6 & \textbf{120.0} & 119.5 & \textbf{129.6} \\
    GPQA-D & 76.8 & \textbf{84.8} & 86.3 & \textbf{96.2} & 94.4 & \textbf{107.1} & 99.4 & \textbf{114.2} & 103.6 & \textbf{121.5} \\
    IFEval & 79.5 & \textbf{92.3} & 93.9 & \textbf{108.8} & 103.4 & \textbf{119.9} & 108.9 & \textbf{128.8} & 117.1 & \textbf{139.7} \\
    HumanEval+ & 128.5 & \textbf{131.8} & 141.9 & \textbf{151.5} & 148.9 & \textbf{160.7} & 143.6 & \textbf{173.8} & 157.6 & \textbf{186.0} \\
    MBPP+ & 121.2 & \textbf{128.4} & 130.1 & \textbf{130.4} & 136.4 & \textbf{149.6} & 142.6 & \textbf{152.9} & 139.0 & \textbf{166.9} \\
    LCB & 87.9 & \textbf{95.3} & 103.7 & \textbf{107.5} & 108.0 & \textbf{119.2} & 113.0 & \textbf{124.1} & 116.2 & \textbf{135.8} \\
    \addlinespace[1pt]
    Average & 103.3 & \textbf{109.3} & 115.5 & \textbf{123.3} & 121.0 & \textbf{135.6} & 126.0 & \textbf{142.0} & 130.6 & \textbf{152.3} \\
    \midrule
    \multicolumn{11}{@{}l}{\textit{TPF}} \\
    GSM8K & 23.69 & \textbf{26.95} & 27.11 & \textbf{30.91} & 28.74 & \textbf{33.69} & 30.04 & \textbf{36.44} & 31.07 & \textbf{38.72} \\
    MATH500 & 25.24 & \textbf{27.37} & 27.57 & \textbf{31.25} & 28.69 & \textbf{34.16} & 29.76 & \textbf{36.69} & 30.65 & \textbf{39.65} \\
    AIME26 & 18.38 & \textbf{19.59} & 21.11 & \textbf{22.76} & 20.61 & \textbf{25.48} & 22.92 & \textbf{25.69} & 23.86 & \textbf{27.83} \\
    GPQA-D & 15.47 & \textbf{17.82} & 17.63 & \textbf{20.42} & 19.18 & \textbf{22.88} & 20.14 & \textbf{24.66} & 21.18 & \textbf{27.78} \\
    IFEval & 16.14 & \textbf{19.97} & 19.22 & \textbf{23.66} & 21.08 & \textbf{26.61} & 22.47 & \textbf{28.71} & 24.11 & \textbf{31.69} \\
    HumanEval+ & 27.02 & \textbf{29.92} & 30.01 & \textbf{34.39} & 30.77 & \textbf{37.53} & 32.18 & \textbf{40.59} & 33.39 & \textbf{43.65} \\
    MBPP+ & 24.71 & \textbf{27.48} & 27.03 & \textbf{30.16} & 28.39 & \textbf{33.40} & 29.75 & \textbf{35.79} & 30.64 & \textbf{38.20} \\
    LCB & 18.12 & \textbf{20.04} & 20.95 & \textbf{23.24} & 22.11 & \textbf{25.59} & 23.05 & \textbf{27.90} & 23.80 & \textbf{30.46} \\
    \addlinespace[1pt]
    Average & 21.10 & \textbf{23.64} & 23.83 & \textbf{27.10} & 24.95 & \textbf{29.92} & 26.29 & \textbf{32.06} & 27.34 & \textbf{34.75} \\
    \midrule
    \multicolumn{11}{@{}l}{\textit{Accuracy}} \\
    GSM8K & 96.4 & \textbf{96.4} & 96.1 & \textbf{95.5} & 95.8 & \textbf{96.1} & 96.0 & \textbf{95.8} & 94.9 & \textbf{95.0} \\
    MATH500 & 96.8 & \textbf{96.4} & 95.8 & \textbf{95.6} & 95.8 & \textbf{92.4} & 95.0 & \textbf{91.2} & 94.8 & \textbf{89.8} \\
    AIME26 & 63.3 & \textbf{76.7} & 60.0 & \textbf{66.7} & 73.3 & \textbf{76.7} & 53.3 & \textbf{56.7} & 60.0 & \textbf{40.0} \\
    GPQA-D & 69.7 & \textbf{68.2} & 71.2 & \textbf{71.7} & 73.7 & \textbf{71.7} & 70.2 & \textbf{71.7} & 74.2 & \textbf{70.2} \\
    IFEval & 96.4 & \textbf{96.4} & 97.0 & \textbf{96.4} & 95.8 & \textbf{95.4} & 95.6 & \textbf{95.1} & 95.9 & \textbf{93.8} \\
    HumanEval+ & 92.1 & \textbf{93.3} & 90.2 & \textbf{92.7} & 91.5 & \textbf{89.6} & 91.5 & \textbf{89.0} & 84.1 & \textbf{82.3} \\
    MBPP+ & 80.4 & \textbf{79.4} & 82.3 & \textbf{81.0} & 80.7 & \textbf{82.3} & 81.7 & \textbf{79.1} & 79.6 & \textbf{79.6} \\
    LCB & 70.5 & \textbf{71.3} & 69.3 & \textbf{68.1} & 70.5 & \textbf{69.3} & 69.6 & \textbf{68.4} & 65.2 & \textbf{62.6} \\
    \addlinespace[1pt]
    Average & 83.2 & \textbf{84.8} & 82.7 & \textbf{83.5} & 84.6 & \textbf{84.2} & 81.6 & \textbf{80.9} & 81.1 & \textbf{76.7} \\
    \bottomrule
  \end{tabular*}
  \caption{DiffusionGemma 26B, per-benchmark results for \Cref{fig:pareto}.}
  \label{tab:appendix_pareto_gemma}
\end{table*}

\begin{table*}[t]
  \centering
  \footnotesize
  \setlength{\tabcolsep}{0pt}
  \begin{tabular*}{\textwidth}{@{\extracolsep{\fill}}lrrrrrrrrrr@{}}
    \toprule
    Confidence threshold & \multicolumn{2}{c}{0.3} & \multicolumn{2}{c}{0.5} & \multicolumn{2}{c}{0.7} & \multicolumn{2}{c}{0.85} & \multicolumn{2}{c}{0.95} \\
    \cmidrule(lr){2-3} \cmidrule(lr){4-5} \cmidrule(lr){6-7} \cmidrule(lr){8-9} \cmidrule(lr){10-11}
    Method & Base & \textbf{\methodname} & Base & \textbf{\methodname} & Base & \textbf{\methodname} & Base & \textbf{\methodname} & Base & \textbf{\methodname} \\
    \midrule
    \multicolumn{11}{@{}l}{\textit{TPS}} \\
    GSM8K & 128.1 & \textbf{146.1} & 133.5 & \textbf{150.1} & 114.2 & \textbf{124.8} & 84.5 & \textbf{91.9} & 58.6 & \textbf{61.5} \\
    MATH500 & 160.6 & \textbf{186.5} & 171.2 & \textbf{199.5} & 160.4 & \textbf{186.1} & 126.5 & \textbf{139.3} & 95.2 & \textbf{102.9} \\
    AIME26 & 109.6 & \textbf{115.8} & 131.3 & \textbf{149.9} & 113.7 & \textbf{125.9} & 92.6 & \textbf{92.9} & 70.6 & \textbf{72.1} \\
    GPQA-D & 65.6 & \textbf{66.2} & 79.8 & \textbf{87.1} & 69.5 & \textbf{73.8} & 54.4 & \textbf{56.1} & 42.8 & \textbf{44.4} \\
    IFEval & 46.5 & \textbf{49.5} & 37.6 & \textbf{40.0} & 31.1 & \textbf{32.4} & 26.8 & \textbf{27.8} & 23.8 & \textbf{24.0} \\
    HumanEval+ & 271.7 & \textbf{345.3} & 271.3 & \textbf{327.6} & 247.2 & \textbf{301.4} & 213.9 & \textbf{251.5} & 168.5 & \textbf{192.7} \\
    MBPP+ & 184.2 & \textbf{218.0} & 183.7 & \textbf{214.9} & 163.6 & \textbf{194.2} & 124.4 & \textbf{148.5} & 91.2 & \textbf{99.3} \\
    LCB & 136.2 & \textbf{157.4} & 137.8 & \textbf{153.0} & 113.7 & \textbf{131.8} & 93.2 & \textbf{101.0} & 68.6 & \textbf{73.3} \\
    \addlinespace[1pt]
    Average & 137.8 & \textbf{160.6} & 143.3 & \textbf{165.3} & 126.7 & \textbf{146.3} & 102.0 & \textbf{113.6} & 77.4 & \textbf{83.8} \\
    \midrule
    \multicolumn{11}{@{}l}{\textit{TPF}} \\
    GSM8K & 5.61 & \textbf{6.42} & 5.74 & \textbf{6.58} & 4.88 & \textbf{5.49} & 3.65 & \textbf{3.98} & 2.55 & \textbf{2.69} \\
    MATH500 & 6.90 & \textbf{8.27} & 7.36 & \textbf{8.78} & 6.85 & \textbf{8.14} & 5.41 & \textbf{6.22} & 4.08 & \textbf{4.59} \\
    AIME26 & 4.62 & \textbf{5.04} & 5.68 & \textbf{6.59} & 4.92 & \textbf{5.46} & 3.88 & \textbf{4.12} & 3.05 & \textbf{3.12} \\
    GPQA-D & 2.84 & \textbf{2.90} & 3.46 & \textbf{3.85} & 3.02 & \textbf{3.26} & 2.36 & \textbf{2.49} & 1.86 & \textbf{1.93} \\
    IFEval & 2.03 & \textbf{2.21} & 1.66 & \textbf{1.77} & 1.35 & \textbf{1.40} & 1.15 & \textbf{1.19} & 1.02 & \textbf{1.04} \\
    HumanEval+ & 12.36 & \textbf{15.60} & 12.01 & \textbf{14.97} & 10.94 & \textbf{13.69} & 9.32 & \textbf{11.38} & 7.40 & \textbf{8.85} \\
    MBPP+ & 7.88 & \textbf{9.63} & 8.04 & \textbf{9.52} & 7.16 & \textbf{8.57} & 5.55 & \textbf{6.54} & 3.96 & \textbf{4.43} \\
    LCB & 5.96 & \textbf{7.02} & 6.09 & \textbf{6.83} & 5.10 & \textbf{5.83} & 4.01 & \textbf{4.47} & 3.03 & \textbf{3.24} \\
    \addlinespace[1pt]
    Average & 6.02 & \textbf{7.14} & 6.26 & \textbf{7.36} & 5.53 & \textbf{6.48} & 4.42 & \textbf{5.05} & 3.37 & \textbf{3.74} \\
    \midrule
    \multicolumn{11}{@{}l}{\textit{Accuracy}} \\
    GSM8K & 81.6 & \textbf{79.9} & 88.6 & \textbf{87.6} & 90.9 & \textbf{89.8} & 91.5 & \textbf{91.1} & 91.3 & \textbf{91.1} \\
    MATH500 & 85.8 & \textbf{84.8} & 89.6 & \textbf{87.6} & 92.8 & \textbf{91.8} & 91.2 & \textbf{93.2} & 92.6 & \textbf{93.2} \\
    AIME26 & 16.7 & \textbf{20.0} & 26.7 & \textbf{30.0} & 46.7 & \textbf{50.0} & 46.7 & \textbf{40.0} & 43.3 & \textbf{53.3} \\
    GPQA-D & 37.4 & \textbf{34.3} & 40.9 & \textbf{46.5} & 52.5 & \textbf{52.0} & 51.0 & \textbf{50.5} & 47.5 & \textbf{49.0} \\
    IFEval & 84.1 & \textbf{83.1} & 86.6 & \textbf{85.6} & 86.5 & \textbf{86.0} & 87.1 & \textbf{87.3} & 86.7 & \textbf{86.9} \\
    HumanEval+ & 64.6 & \textbf{67.1} & 78.0 & \textbf{72.0} & 83.5 & \textbf{81.7} & 85.4 & \textbf{83.5} & 85.4 & \textbf{84.8} \\
    MBPP+ & 57.4 & \textbf{57.7} & 66.4 & \textbf{62.4} & 69.8 & \textbf{69.0} & 70.4 & \textbf{73.0} & 73.5 & \textbf{73.0} \\
    LCB & 17.0 & \textbf{14.3} & 20.2 & \textbf{20.5} & 25.7 & \textbf{26.6} & 28.1 & \textbf{28.1} & 29.2 & \textbf{30.4} \\
    \addlinespace[1pt]
    Average & 55.6 & \textbf{55.2} & 62.1 & \textbf{61.5} & 68.6 & \textbf{68.4} & 68.9 & \textbf{68.3} & 68.7 & \textbf{70.2} \\
    \bottomrule
  \end{tabular*}
  \caption{LLaDA 2.1 mini 16B, per-benchmark results for \cref{fig:pareto}.}
  \label{tab:appendix_pareto_llada}
\end{table*}

\begin{table}[t]
  \centering
  \footnotesize
  \setlength{\tabcolsep}{0pt}
  \begin{tabular*}{\columnwidth}{@{\extracolsep{\fill}}lrrrrr@{}}
    \toprule
    Block & 128 & 256 & \textbf{256} & 384 & 512 \\
     & & & \textbf{\methodname} & & \\
    \midrule
    \multicolumn{6}{@{}l}{\textit{TPS}} \\
    GSM8K & 90.9 & 136.1 & \textbf{150.8} & 171.0 & 177.8 \\
    MATH500 & 96.6 & 139.9 & \textbf{156.0} & 167.8 & 171.1 \\
    AIME26 & 86.7 & 101.0 & \textbf{121.6} & 118.9 & 112.6 \\
    GPQA-D & 70.4 & 94.4 & \textbf{107.1} & 106.0 & 104.8 \\
    IFEval & 66.1 & 103.4 & \textbf{119.9} & 123.0 & 135.9 \\
    HumanEval+ & 99.9 & 148.9 & \textbf{160.7} & 172.3 & 188.8 \\
    MBPP+ & 90.5 & 136.4 & \textbf{149.6} & 165.2 & 183.7 \\
    LCB & 78.1 & 108.0 & \textbf{119.2} & 118.7 & 118.8 \\
    \addlinespace[1pt]
    Average & 84.9 & 121.0 & \textbf{135.6} & 142.9 & 149.2 \\
    \midrule
    \multicolumn{6}{@{}l}{\textit{TPF}} \\
    GSM8K & 18.59 & 28.74 & \textbf{33.69} & 36.62 & 40.96 \\
    MATH500 & 19.47 & 28.69 & \textbf{34.16} & 35.15 & 38.15 \\
    AIME26 & 16.43 & 20.61 & \textbf{25.48} & 24.55 & 24.70 \\
    GPQA-D & 13.43 & 19.18 & \textbf{22.88} & 22.04 & 22.95 \\
    IFEval & 13.53 & 21.08 & \textbf{26.61} & 26.67 & 30.54 \\
    HumanEval+ & 20.15 & 30.77 & \textbf{37.53} & 38.34 & 42.98 \\
    MBPP+ & 18.01 & 28.39 & \textbf{33.40} & 35.66 & 40.65 \\
    LCB & 15.41 & 22.11 & \textbf{25.59} & 25.04 & 26.12 \\
    \addlinespace[1pt]
    Average & 16.88 & 24.95 & \textbf{29.92} & 30.51 & 33.38 \\
    \midrule
    \multicolumn{6}{@{}l}{\textit{Accuracy}} \\
    GSM8K & 95.5 & 95.8 & \textbf{96.1} & 95.8 & 95.5 \\
    MATH500 & 96.4 & 95.8 & \textbf{92.4} & 94.6 & 93.2 \\
    AIME26 & 73.3 & 73.3 & \textbf{76.7} & 43.3 & 43.3 \\
    GPQA-D & 73.2 & 73.7 & \textbf{71.7} & 73.7 & 65.7 \\
    IFEval & 96.3 & 95.8 & \textbf{95.4} & 96.2 & 93.9 \\
    HumanEval+ & 94.5 & 91.5 & \textbf{89.6} & 90.9 & 91.5 \\
    MBPP+ & 81.7 & 80.7 & \textbf{82.3} & 79.1 & 80.4 \\
    LCB & 71.1 & 70.5 & \textbf{69.3} & 68.1 & 62.0 \\
    \addlinespace[1pt]
    Average & 85.3 & 84.6 & \textbf{84.2} & 80.2 & 78.2 \\
    \bottomrule
  \end{tabular*}
  \caption{DiffusionGemma 26B, per-benchmark results for \Cref{fig:ablation_block}.}
  \label{tab:appendix_block_gemma}
\end{table}

\begin{table}[t]
  \centering
  \footnotesize
  \setlength{\tabcolsep}{0pt}
  \begin{tabular*}{\columnwidth}{@{\extracolsep{\fill}}lrrrrrr@{}}
    \toprule
    Block & 32 & \textbf{32} & 64 & \textbf{64} & 96 & 128 \\
     & & \textbf{\methodname} & & \textbf{\methodname} & & \\
    \midrule
    \multicolumn{7}{@{}l}{\textit{TPS}} \\
    GSM8K & 101.5 & \textbf{122.8} & 114.2 & \textbf{124.8} & 113.5 & 91.7 \\
    MATH500 & 133.9 & \textbf{163.9} & 160.4 & \textbf{186.1} & 159.7 & 148.5 \\
    AIME26 & 103.6 & \textbf{118.5} & 113.7 & \textbf{125.9} & 121.3 & 105.4 \\
    GPQA-D & 65.8 & \textbf{73.4} & 69.5 & \textbf{73.8} & 70.5 & 62.9 \\
    IFEval & 32.8 & \textbf{35.4} & 31.1 & \textbf{32.4} & 28.8 & 25.3 \\
    HumanEval+ & 204.2 & \textbf{257.0} & 247.2 & \textbf{301.4} & 249.6 & 188.4 \\
    MBPP+ & 142.3 & \textbf{173.4} & 163.6 & \textbf{194.2} & 168.3 & 138.6 \\
    LCB & 105.1 & \textbf{122.3} & 113.7 & \textbf{131.8} & 115.1 & 93.0 \\
    \addlinespace[1pt]
    Average & 111.1 & \textbf{133.3} & 126.7 & \textbf{146.3} & 128.3 & 106.7 \\
    \midrule
    \multicolumn{7}{@{}l}{\textit{TPF}} \\
    GSM8K & 4.36 & \textbf{5.32} & 4.88 & \textbf{5.49} & 4.86 & 4.05 \\
    MATH500 & 5.82 & \textbf{7.16} & 6.85 & \textbf{8.14} & 7.02 & 6.45 \\
    AIME26 & 4.41 & \textbf{5.12} & 4.92 & \textbf{5.46} & 5.10 & 4.46 \\
    GPQA-D & 2.80 & \textbf{3.18} & 3.02 & \textbf{3.26} & 3.03 & 2.71 \\
    IFEval & 1.41 & \textbf{1.52} & 1.35 & \textbf{1.40} & 1.24 & 1.09 \\
    HumanEval+ & 9.14 & \textbf{11.52} & 10.94 & \textbf{13.69} & 10.93 & 8.33 \\
    MBPP+ & 6.14 & \textbf{7.73} & 7.16 & \textbf{8.57} & 7.37 & 6.16 \\
    LCB & 4.53 & \textbf{5.45} & 5.10 & \textbf{5.83} & 5.02 & 4.17 \\
    \addlinespace[1pt]
    Average & 4.83 & \textbf{5.88} & 5.53 & \textbf{6.48} & 5.57 & 4.68 \\
    \midrule
    \multicolumn{7}{@{}l}{\textit{Accuracy}} \\
    GSM8K & 90.1 & \textbf{90.4} & 90.9 & \textbf{89.8} & 89.7 & 80.7 \\
    MATH500 & 91.6 & \textbf{92.4} & 92.8 & \textbf{91.8} & 90.2 & 85.8 \\
    AIME26 & 46.7 & \textbf{56.7} & 46.7 & \textbf{50.0} & 46.7 & 30.0 \\
    GPQA-D & 52.5 & \textbf{50.5} & 52.5 & \textbf{52.0} & 46.5 & 45.5 \\
    IFEval & 90.6 & \textbf{89.3} & 86.5 & \textbf{86.0} & 86.2 & 78.3 \\
    HumanEval+ & 82.9 & \textbf{81.7} & 83.5 & \textbf{81.7} & 75.6 & 61.0 \\
    MBPP+ & 70.4 & \textbf{69.6} & 69.8 & \textbf{69.0} & 68.8 & 70.9 \\
    LCB & 25.7 & \textbf{26.0} & 25.7 & \textbf{26.6} & 24.6 & 20.5 \\
    \addlinespace[1pt]
    Average & 68.8 & \textbf{69.6} & 68.6 & \textbf{68.4} & 66.0 & 59.1 \\
    \bottomrule
  \end{tabular*}
  \caption{LLaDA 2.1 mini 16B, per-benchmark results for \Cref{fig:ablation_block}.}
  \label{tab:appendix_block_llada}
\end{table}

\begin{table*}[t]
  \centering
  \footnotesize
  \setlength{\tabcolsep}{0pt}
  \begin{tabular*}{\textwidth}{@{\extracolsep{\fill}}lrrrrrrrr@{}}
    \toprule
    Block & \multicolumn{4}{c}{32} & \multicolumn{4}{c}{64} \\
    \cmidrule(lr){2-5} \cmidrule(lr){6-9}
    Method & Base & FlowBlock & MBD-LMs & \textbf{\methodname} & Base & FlowBlock & MBD-LMs & \textbf{\methodname} \\
    \midrule
    \multicolumn{9}{@{}l}{\textit{TPS}} \\
    GSM8K & 101.5 & \textbf{153.2} & 119.9 & 122.8 & 114.2 & \textbf{153.8} & 119.3 & 124.8 \\
    MATH500 & 133.9 & \textbf{207.1} & 161.8 & 163.9 & 160.4 & \textbf{210.8} & 176.6 & 186.1 \\
    AIME26 & 103.6 & \textbf{143.6} & 116.4 & 118.5 & 113.7 & \textbf{147.7} & 114.0 & 125.9 \\
    GPQA-D & 65.8 & \textbf{91.1} & 74.2 & 73.4 & 69.5 & \textbf{83.4} & 69.7 & 73.8 \\
    IFEval & 32.8 & 26.5 & 35.1 & \textbf{35.4} & 31.1 & 24.1 & 31.4 & \textbf{32.4} \\
    HumanEval+ & 204.2 & 227.9 & 253.3 & \textbf{257.0} & 247.2 & 236.0 & 299.7 & \textbf{301.4} \\
    MBPP+ & 142.3 & 118.7 & 167.8 & \textbf{173.4} & 163.6 & 116.9 & \textbf{195.5} & 194.2 \\
    LCB & 105.1 & 85.1 & 120.5 & \textbf{122.3} & 113.7 & 86.0 & 128.6 & \textbf{131.8} \\
    \addlinespace[1pt]
    Average & 111.1 & 131.7 & 131.1 & \textbf{133.3} & 126.7 & 132.3 & 141.8 & \textbf{146.3} \\
    \midrule
    \multicolumn{9}{@{}l}{\textit{TPF}} \\
    GSM8K & 4.36 & \textbf{6.66} & 5.32 & 5.32 & 4.88 & \textbf{6.74} & 5.33 & 5.49 \\
    MATH500 & 5.82 & \textbf{9.06} & 7.32 & 7.16 & 6.85 & \textbf{9.55} & 8.03 & 8.14 \\
    AIME26 & 4.41 & \textbf{6.27} & 5.18 & 5.12 & 4.92 & \textbf{6.51} & 5.06 & 5.46 \\
    GPQA-D & 2.80 & \textbf{3.92} & 3.27 & 3.18 & 3.02 & \textbf{3.67} & 3.14 & 3.26 \\
    IFEval & 1.41 & 1.15 & \textbf{1.55} & 1.52 & 1.35 & 1.04 & 1.38 & \textbf{1.40} \\
    HumanEval+ & 9.14 & 10.24 & \textbf{11.72} & 11.52 & 10.94 & 10.62 & \textbf{13.70} & 13.69 \\
    MBPP+ & 6.14 & 5.20 & \textbf{7.74} & 7.73 & 7.16 & 5.24 & \textbf{8.69} & 8.57 \\
    LCB & 4.53 & 3.67 & \textbf{5.46} & 5.45 & 5.10 & 3.78 & \textbf{5.87} & 5.83 \\
    \addlinespace[1pt]
    Average & 4.83 & 5.77 & \textbf{5.95} & 5.88 & 5.53 & 5.89 & 6.40 & \textbf{6.48} \\
    \midrule
    \multicolumn{9}{@{}l}{\textit{Accuracy}} \\
    GSM8K & 90.1 & \textbf{90.4} & \textbf{90.4} & \textbf{90.4} & 90.9 & \textbf{91.1} & 88.2 & 89.8 \\
    MATH500 & 91.6 & 91.8 & 90.0 & \textbf{92.4} & 92.8 & 91.6 & 88.6 & \textbf{91.8} \\
    AIME26 & 46.7 & 36.7 & 40.0 & \textbf{56.7} & 46.7 & 40.0 & 40.0 & \textbf{50.0} \\
    GPQA-D & 52.5 & 49.0 & 50.0 & \textbf{50.5} & 52.5 & 50.0 & 41.4 & \textbf{52.0} \\
    IFEval & 90.6 & \textbf{90.5} & 89.3 & 89.3 & 86.5 & 85.1 & 84.4 & \textbf{86.0} \\
    HumanEval+ & 82.9 & \textbf{84.1} & 82.9 & 81.7 & 83.5 & \textbf{84.1} & 79.9 & 81.7 \\
    MBPP+ & 70.4 & \textbf{72.0} & 71.2 & 69.6 & 69.8 & \textbf{74.6} & 72.8 & 69.0 \\
    LCB & 25.7 & \textbf{28.4} & 25.4 & 26.0 & 25.7 & \textbf{29.8} & 27.5 & 26.6 \\
    \addlinespace[1pt]
    Average & 68.8 & 67.9 & 67.4 & \textbf{69.6} & 68.6 & 68.3 & 65.3 & \textbf{68.4} \\
    \bottomrule
  \end{tabular*}
  \caption{LLaDA 2.1 mini 16B, per-benchmark results for \Cref{tab:baselines_llada}. Bold marks the best multi-block method per row and block size.}
  \label{tab:appendix_baselines_llada}
\end{table*}

\begin{table}[t]

  \centering
  \footnotesize
  \setlength{\tabcolsep}{0pt}
  \begin{tabular*}{\columnwidth}{@{\extracolsep{\fill}}lrr@{}}
    \toprule
    Benchmark & DiffusionGemma, 3 seeds & LLaDA 2.1 mini \\
    \midrule
    GSM8K      & $+0.3$ [$-0.1$, $+0.7$]   & $-1.1$ [$-2.3$, $+0.2$] \\
    MATH500    & $-2.3$ [$-4.7$, $+0.2$]   & $-1.0$ [$-3.6$, $+1.4$] \\
    AIME26     & $+1.1$ [$-16.1$, $+18.3$] & $+3.3$ [$-16.7$, $+26.7$] \\
    GPQA-D     & $+0.7$ [$-5.3$, $+6.6$]   & $-0.5$ [$-7.6$, $+6.6$] \\
    IFEval     & $-0.7$ [$-3.1$, $+1.6$]   & $-0.5$ [$-1.7$, $+0.8$] \\
    HumanEval+ & $-2.4$ [$-9.4$, $+4.5$]   & $-1.8$ [$-6.7$, $+3.0$] \\
    MBPP+      & $+0.4$ [$-2.4$, $+3.1$]   & $-0.8$ [$-3.4$, $+1.9$] \\
    LCB        & $+0.6$ [$-7.0$, $+8.1$]   & $+0.9$ [$-2.3$, $+4.1$] \\
    \addlinespace[1pt]
    Average    & $-0.3$ [$-2.1$, $+1.5$]   & $-0.2$ [$-3.1$, $+2.8$] \\
    \bottomrule
  \end{tabular*}
  \caption{Accuracy change of \methodname{} over the base model ($\Delta pp$) with 95\% intervals on the eight light benchmarks. DiffusionGemma samples, so it is run with seeds 11, 12 and 1234. LLaDA 2.1 mini decodes greedily, so its intervals come from a paired bootstrap over problems.}
  \label{tab:appendix_uncertainty}

\end{table}

\section{Block Boundary Inefficiency by Block Index and Dataset}
\label[appendix]{sec:appendix_block}

\begin{figure*}[p]
  \centering
  \includegraphics{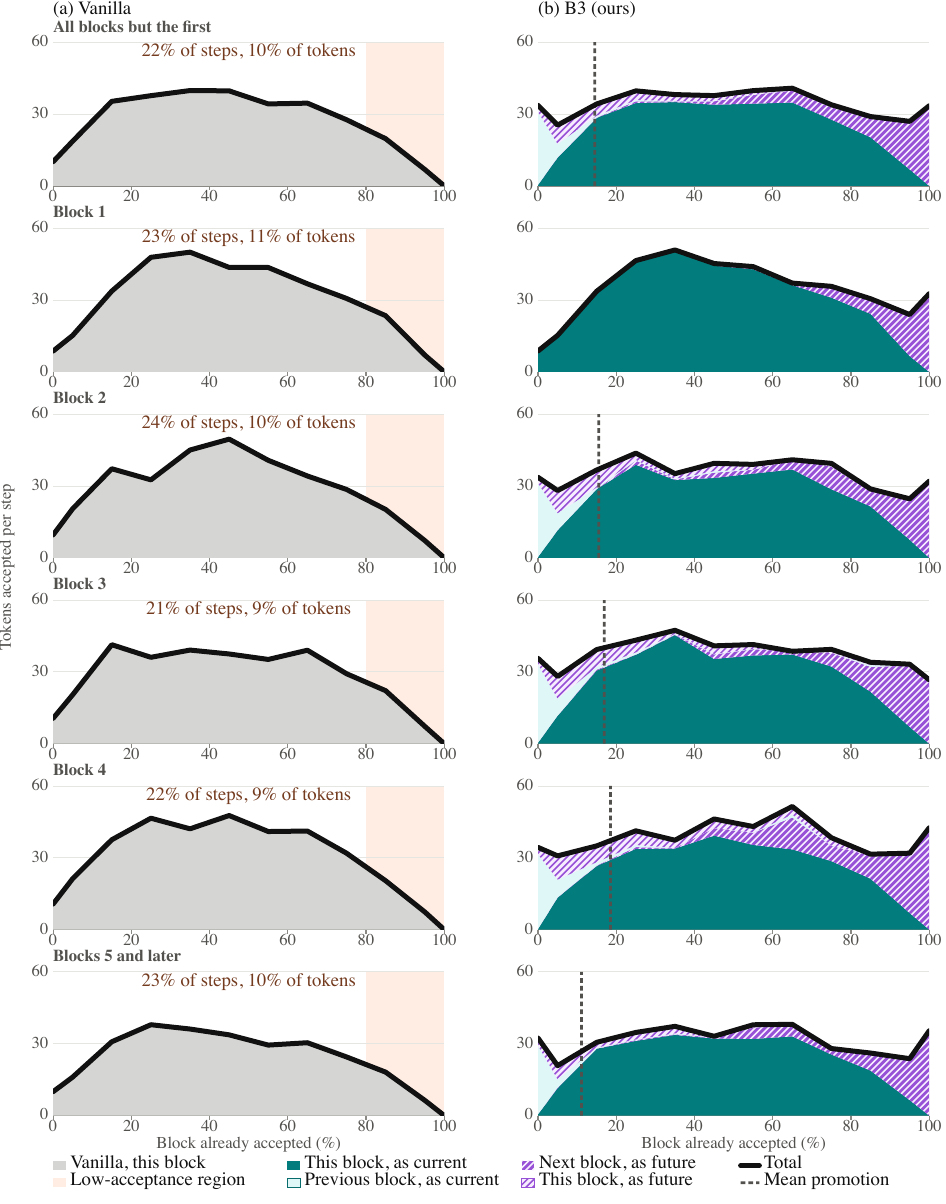}
  \caption{DiffusionGemma on HumanEval+: tokens accepted per step by block index. Vanilla left, \methodname right.}
  \label{fig:splits_gemma_humaneval}
\end{figure*}

\begin{figure*}[p]
  \centering
  \includegraphics{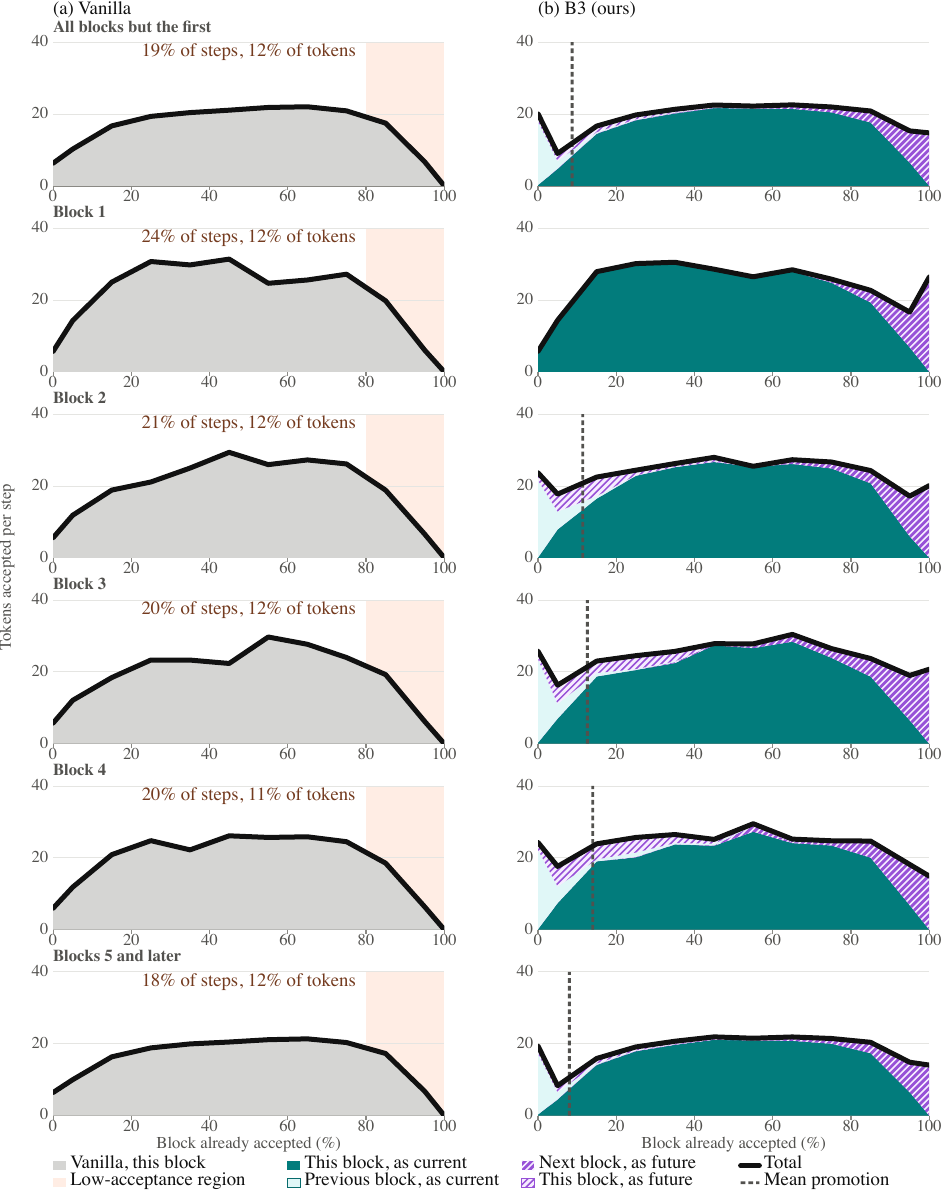}
  \caption{DiffusionGemma on GPQA Diamond: tokens accepted per step by block index. Vanilla left, \methodname right.}
  \label{fig:splits_gemma_gpqa}
\end{figure*}

\Cref{fig:splits_gemma_humaneval,fig:splits_gemma_gpqa} break \Cref{fig:teaser} down by block index for DiffusionGemma on HumanEval+ and GPQA Diamond. Each row plots the tokens accepted per denoising step against the share of the block already accepted, for vanilla decoding~(a, left) and for \methodname~(b, right), on a scale shared by all rows. The shaded region is the block's last 20\%: the steps taken while more than 80\% of the block is accepted. Its label gives the share of vanilla's steps that fall there and the share of tokens those steps accept. At every block index, vanilla spends 19--24\% of its steps there for 10--13\% of the tokens. The last block of each generation is excluded, and the first row also excludes the first block, as in \Cref{fig:teaser}.

\begin{figure*}[t]
  \centering
  \includegraphics{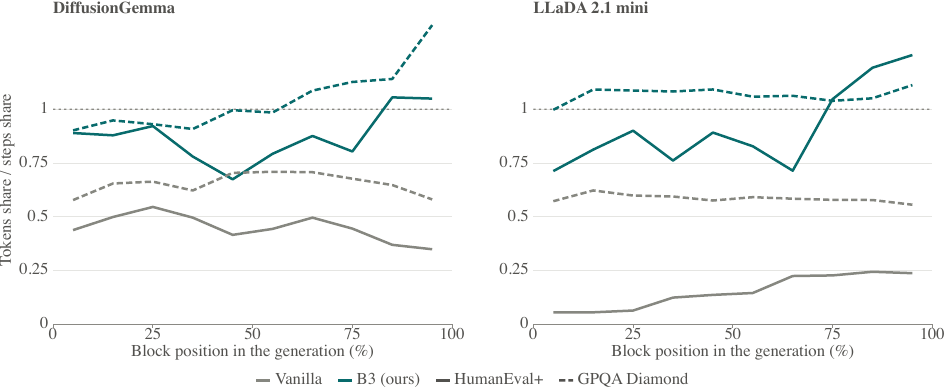}
  \caption{Share of a block's tokens accepted in its last 20\% (more than 80\% of the block accepted), divided by the share of its denoising steps taken there, by the block's position in the generation (first block 0\%, last 100\%; 10 bins). Below 1: the end of the block yields fewer tokens than the steps it costs. Solid: HumanEval+; dashed: GPQA Diamond. Dotted: parity.}
  \label{fig:end_ratio}
\end{figure*}

\Cref{fig:end_ratio} summarises this statistic for both models: the share of a block's tokens accepted in its last 20\%, divided by the share of its steps spent there, by the block's position in the generation. A value of 1 means the end of the block yields tokens in proportion to the steps it costs. Vanilla decoding stays below 1 throughout, at 0.35--0.71 for DiffusionGemma and 0.05--0.62 for LLaDA 2.1 mini. With \methodname{} the ratio is 0.67--1.39 and 0.71--1.25, since the next block accepts tokens on the same forward passes.

LLaDA 2.1 mini on HumanEval+ has the lowest ratio because LLaDA 2.1 forces editing steps once a block is fully accepted, and these accept no new tokens: 39\% of its first block's steps run on a fully accepted block.

\section{Spawn Points of the Self-Calibrating Gate}
\label[appendix]{sec:appendix_spawn}

\begin{figure*}[tp]
  \centering
  \includegraphics{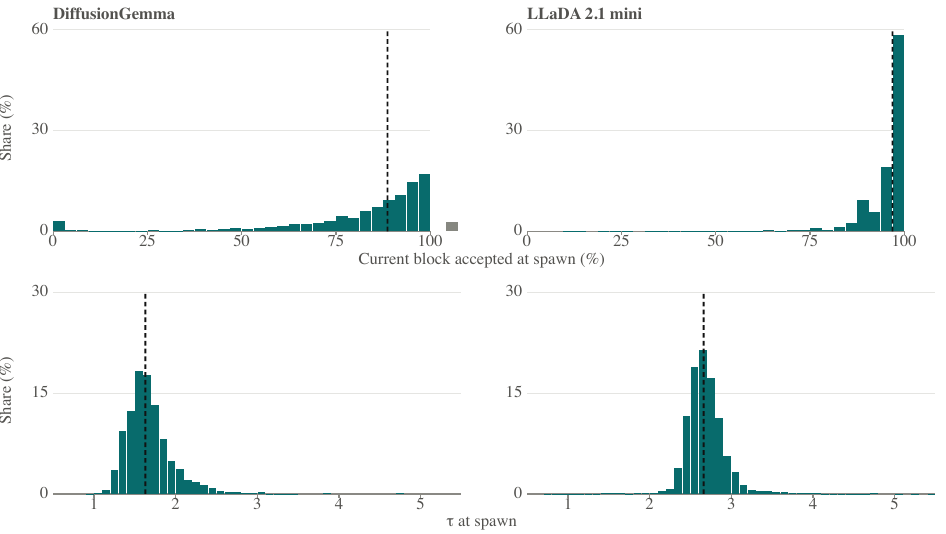}
  \caption{Spawn points of \methodname, pooled over HumanEval+ and GPQA Diamond. Top: share of the current block accepted when the next block spawns; gray: blocks that finished without a spawn. Bottom: $\tau$ at spawn. Dashed: median.}
  \label{fig:spawn_overall}
\end{figure*}

\begin{figure*}[tp]
  \centering
  \includegraphics{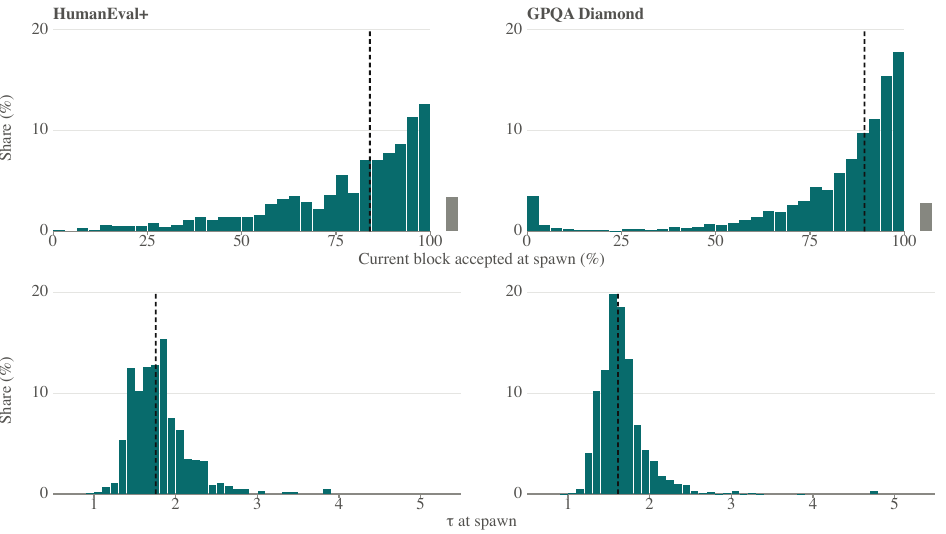}
  \caption{DiffusionGemma: spawn points per dataset. Top: share of the current block accepted at spawn; gray: blocks that finished without a spawn. Bottom: $\tau$ at spawn. Dashed: median.}
  \label{fig:spawn_dataset_gemma}
\end{figure*}

\begin{figure*}[tp]
  \centering
  \includegraphics{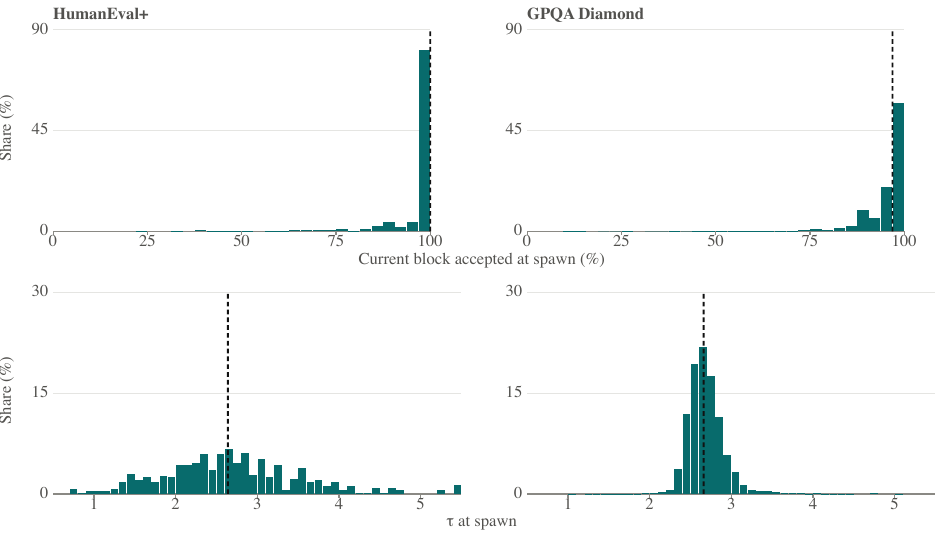}
  \caption{LLaDA 2.1 mini: spawn points per dataset. Top: share of the current block accepted at spawn; gray: blocks that finished without a spawn. Bottom: $\tau$ at spawn. Dashed: median.}
  \label{fig:spawn_dataset_llada}
\end{figure*}

\begin{figure*}[tp]
  \centering
  \includegraphics{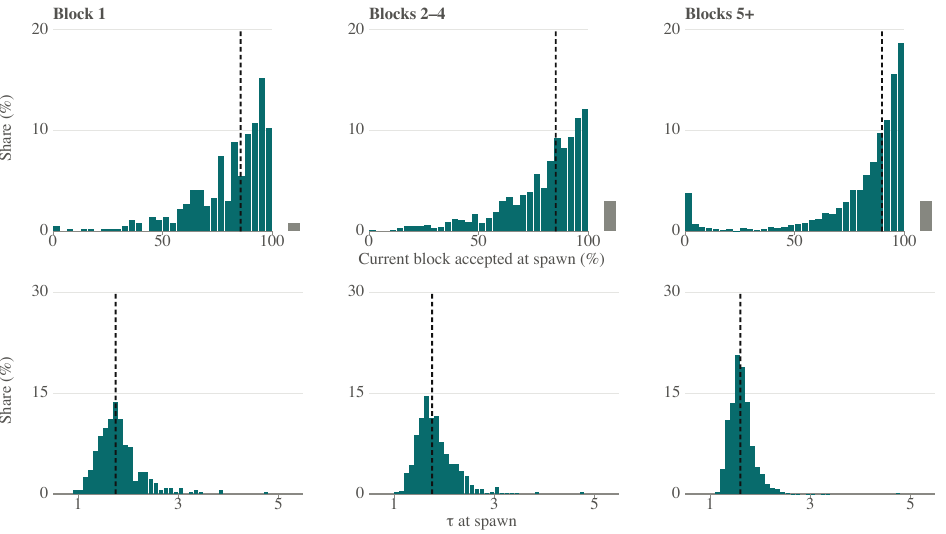}
  \caption{DiffusionGemma: spawn points per block index, pooled over datasets. Top: share of the current block accepted at spawn; gray: blocks that finished without a spawn. Bottom: $\tau$ at spawn. Dashed: median.}
  \label{fig:spawn_index_gemma}
\end{figure*}

\begin{figure*}[tp]
  \centering
  \includegraphics{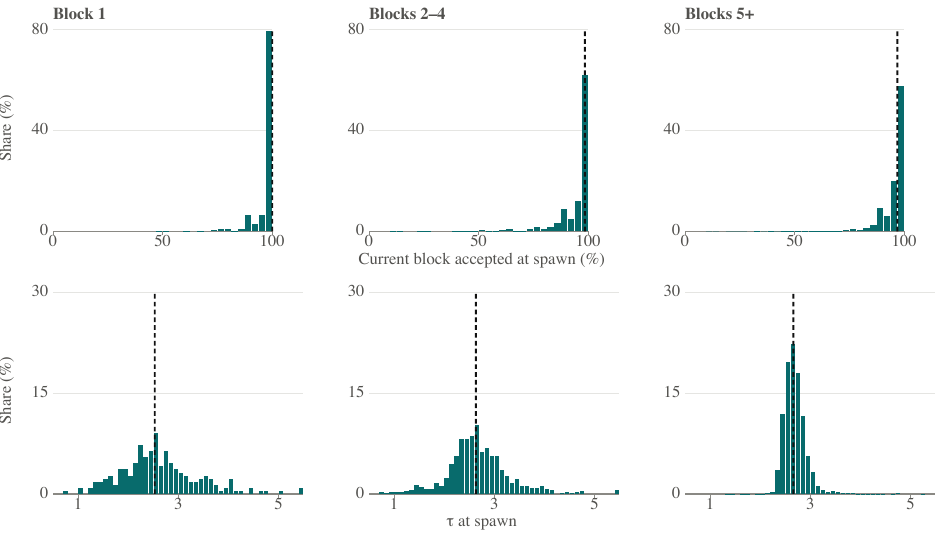}
  \caption{LLaDA 2.1 mini: spawn points per block index, pooled over datasets. Top: share of the current block accepted at spawn; gray: blocks that finished without a spawn. Bottom: $\tau$ at spawn. Dashed: median.}
  \label{fig:spawn_index_llada}
\end{figure*}

\Cref{fig:spawn_overall,fig:spawn_dataset_gemma,fig:spawn_dataset_llada,fig:spawn_index_gemma,fig:spawn_index_llada} show where the gate of \Cref{sec:method_spawning} opens the future block, for DiffusionGemma and LLaDA 2.1 mini on HumanEval+ and GPQA Diamond. For every spawn, the top row gives the share of the current block already accepted, and the bottom row the threshold $\tau$ at that step. Gray bars count blocks that finished without spawning a successor. Spawns at the start of a generation, before $\tau$ has any candidate, have no $\tau$ and are left out of the bottom row.

The spawn point differs per dataset. DiffusionGemma spawns at a median of 84.0\% of the block accepted on HumanEval+ and 89.5\% on GPQA Diamond, at a median $\tau$ of 1.76 and 1.61. Across block indices its medians stay within 85--90\%; blocks 5 and later come mostly from the longer GPQA Diamond generations. LLaDA 2.1 mini spawns close to the end of the block, at a median of 100\% on HumanEval+ and 96.9\% on GPQA Diamond.

\section{Frontier Advance within a Block}
\label[appendix]{sec:appendix_ltr}

\Cref{sec:method_spawning} treats each advance of the frontier $r_i$ as a stand-in for spawning: in \cref{eq:left_gate}, the token at $r_i$ plays the first token of $B_{b+1}$ and the positions $\ell < r_i$ play $B_b$. This requires only two properties of the commitment order within a block:
\begin{enumerate}[label=(\roman*)]
  \item \emph{Gradual advance.} $r_i$ moves through the block in small steps. An early jump to the block's end would yield one candidate over a mostly unsettled block and then freeze $r_i$, so the block would contribute no further candidates.
  \item \emph{Left context drives the frontier.} The token at $r_i$ is accepted mainly on the basis of the partly settled region to its left, as $B_{b+1}$ is conditioned on $B_b$.
\end{enumerate}
Commitment in positional order, one token per step, or with all of $\{\ell < r_i\}$ accepted is not required; \cref{eq:left_gate} takes the maximum over all positions left of $r_i$, as \cref{eq:gate} does over $B_b$. Self-editing behind the frontier is allowed.

\paragraph{Evidence in the literature.}
Confidence- and entropy-based samplers accept the most certain tokens first, inducing a task-dependent, easy-first order~\cite{kim2025ordering}. On text, this order is commonly found to correlate with position. Sumi~\cite{ye2026sumi}, a uniform-state DLM trained from scratch without blocks, shows a left-to-right commitment gradient on three of four tasks under confidence sampling, but not under ancestral sampling, so the gradient emerges at inference rather than being imposed by training. For DiffusionGemma, \citet{asaria2026neither} report a partial left-to-right bias in commitment order on prose, code and math. \citet{hou2026rethinking} find unmasking concentrated near the diagonal both for the block model SDAR and for the full-canvas LLaDA.

\begin{figure*}[t]
  \centering
  \includegraphics{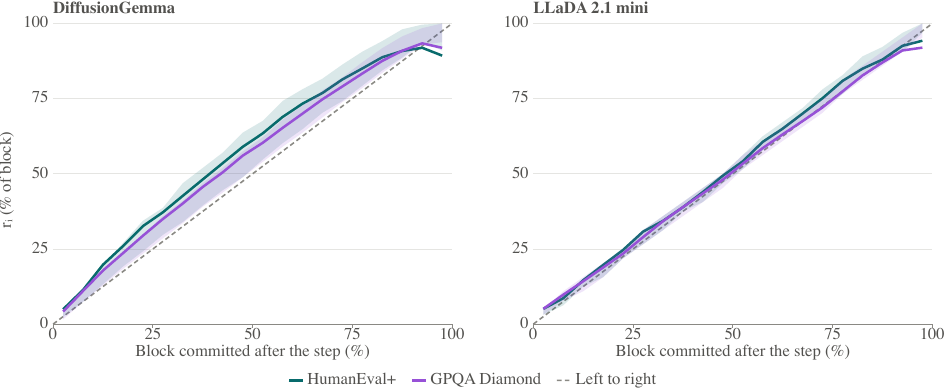}
  \caption{Rightmost position accepted on a denoising step against the share of the block committed after it, vanilla decoding. Lines: mean; bands: P25--P75; dashed: strictly left-to-right.}
  \label{fig:ltr}
\end{figure*}

\paragraph{Measurements on the evaluated models.}
\Cref{fig:ltr} plots $r_i$ against the share of the block committed after step $i$, for DiffusionGemma and LLaDA 2.1 mini under vanilla decoding on HumanEval+ and GPQA Diamond. An early jump of the frontier would lie far above the diagonal. LLaDA 2.1 mini stays within 3.4 points of it, and DiffusionGemma runs ahead by 6--8 points on average and at most 11.4, so in both models $r_i$ advances gradually, satisfying (i). The drop in the last bin comes from steps on an almost complete block, where the remaining acceptances can fall anywhere. The last block of each generation is excluded.

\begin{figure*}[t]
  \centering
  \includegraphics{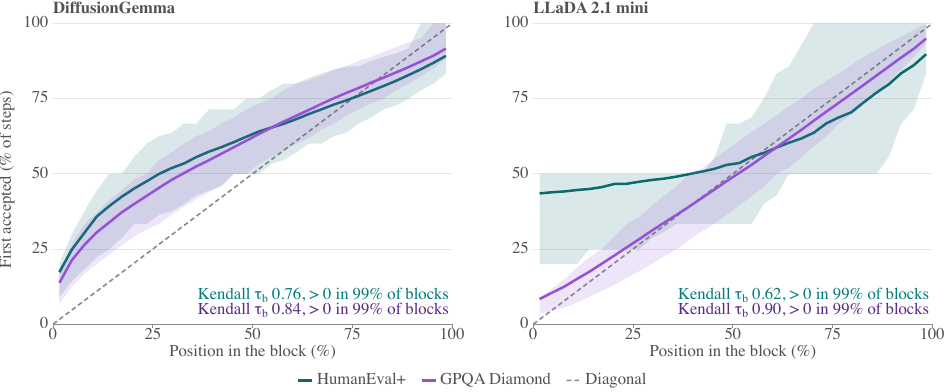}
  \caption{Step at which each position is first accepted, as a share of the block's denoising steps that accept at least one token, against its position in the block, vanilla decoding. Lines: mean; bands: P25--P75; dashed: diagonal. Kendall $\tau_b$: median over blocks of the rank correlation between position and first-acceptance step, and the share of blocks where it is positive.}
  \label{fig:ltr_first}
\end{figure*}

\Cref{fig:ltr_first} plots, for every position in a block, the step of its first acceptance as a share of the block's steps that accept at least one token. Per block, we measure the rank correlation between position and first-acceptance step with Kendall's $\tau_b$~\cite{kendall1938rank,kendall1945ties} (unrelated to the threshold $\tau$), which handles ties from tokens accepted on the same step. Its median is 0.76 and 0.84 for DiffusionGemma on HumanEval+ and GPQA Diamond, and 0.62 and 0.90 for LLaDA 2.1 mini. It is positive in 99\% of blocks in every case, yet clearly below 1 with a wide spread across positions: commitment is correlated with position without being sequential, consistent with (ii). Positions never accepted (0.4\% of DiffusionGemma's) are left out.

\section{Self-Editing of Drafted Tokens}
\label[appendix]{sec:appendix_draft_edits}

The future block is drafted while the current block is still being denoised, so it is conditioned on a prefix that may still hold noise (\Cref{sec:method}). Self-editing is what lets the sampler revise those early guesses once the prefix settles. \Cref{tab:appendix_draft_edits} measures how often this happens. For every token of a block that is finalised, it compares the token accepted first with the token the block finalises with, on all 164 HumanEval+ problems. The vanilla rows count tokens accepted in the current block, the only kind vanilla decoding has; the \methodname{} rows count tokens accepted while their block was still the future block.

Tokens drafted in the future block are edited more often: 1.98\% against 0.27\% for DiffusionGemma, and 4.29\% against 2.80\% for LLaDA 2.1 mini. These are the small mistakes introduced by drafting on an unfinished previous block, and self-editing corrects them before the block is finalised. They stay few: 98.0\% and 95.7\% of the drafted tokens are kept unchanged.

\begin{table}[t]
  \centering
  \footnotesize
  \setlength{\tabcolsep}{3pt}
  \begin{tabular*}{\columnwidth}{@{\extracolsep{\fill}}lrrr@{}}
    \toprule
    Accepted in & Tokens & Edited later & Share \\
    \midrule
    \multicolumn{4}{@{}l}{\textit{DiffusionGemma}} \\
    Vanilla & 237{,}496 & 633 & 0.27\% \\
    \methodname{} (future block) & 42{,}426 & 839 & 1.98\% \\
    \midrule
    \multicolumn{4}{@{}l}{\textit{LLaDA 2.1 mini}} \\
    Vanilla & 51{,}520 & 1{,}441 & 2.80\% \\
    \methodname{} (future block) & 26{,}250 & 1{,}125 & 4.29\% \\
    \bottomrule
  \end{tabular*}
  \caption{Tokens edited after their first acceptance, on HumanEval+.}
  \label{tab:appendix_draft_edits}
\end{table}

\section{DiffusionGemma Notation Clarifications}
\label[appendix]{sec:appendix_gemma}

Two implementation details of DiffusionGemma~\cite{diffusiongemma2026} differ slightly from the general description of block diffusion in \Cref{sec:method_spawning}. Neither changes the notation or the method.

\paragraph{Revoked acceptances.}
In DiffusionGemma, an accepted token can in principle be unaccepted on a later forward pass. Its position is then removed from the set of accepted positions and its token is replaced by noise drawn from $\pi$ for the next forward pass. The set $I_i$ of accepted positions, and hence the frontier $r_i = \max I_i$, are therefore not guaranteed to be monotone in $i$. This is a very rare occurrence, and the definitions of $I_i$, $r_i$ and $C_\tau$ apply unchanged.

\paragraph{Causal history.}
DiffusionGemma encodes its history with fully causal attention, which does not change the forward pass of \methodname{}. Our method operates on DiffusionGemma's token-level causal KV cache, and only the current and future blocks attend block-causally (\Cref{fig:attention}). Once the current block is finalized (\emph{committed}, in DiffusionGemma's terminology), it is re-encoded with token-level causal attention and pushed to the history, exactly as in the regular DiffusionGemma inference stack.

\section{LLM Use in Research}
We used Claude Code~\cite{anthropic2026claudecode} for assistance with code writing and code reviewing, as well as LaTeX writing and figure generation. We also used Claude Chat~\cite{anthropic2026claude} for text drafting and as a reviewer of the paper.

\end{document}